\documentclass[runningheads]{llncs}
\usepackage{graphicx}
\usepackage{xcolor}
\usepackage{comment}
\usepackage{amsmath,amssymb}
\usepackage{color}
\usepackage[accsupp]{axessibility}
\usepackage{booktabs}
\usepackage{tabularx}
\usepackage[inline]{enumitem}
\usepackage{tikz}
\usetikzlibrary{quotes, arrows.meta, angles, calc, 3d, shapes, intersections, plotmarks, positioning, backgrounds, external}
\usepackage[capitalize]{cleveref}

\makeatletter\let\captiontemp\@makecaption\makeatother
\usepackage[font=footnotesize,labelformat=simple]{subcaption}
\makeatletter\let\@makecaption\captiontemp\makeatother

\makeatletter
\renewcommand{\paragraph}{%
  \@startsection{paragraph}{4}%
  {\z@}{0.5em}{-1em}%
  {\normalfont\normalsize\bfseries}%
}
\makeatother

\crefname{section}{Sec.}{Secs.}
\Crefname{section}{Section}{Sections}
\Crefname{table}{Table}{Tables}
\crefname{table}{Tab.}{Tabs.}

\makeatletter
\newcommand{\lambdasym}{\lambda\!\reflectbox{$\lambda$}}
\newcommand{\printfnsymbol}[1]{\textsuperscript{\@fnsymbol{#1}}}
\makeatother
\graphicspath{{./figures/}}
\newcolumntype{C}{>{\centering\arraybackslash}X}

\usepackage{amsmath,amsfonts,bm}

\def\eqref#1{equation~\ref{#1}}

\def\1{\bm{1}}

\def\vtheta{{\bm{\theta}}}

\def\vc{{\bm{c}}}
\def\vd{{\bm{d}}}

\def\vf{{\bm{f}}}

\def\vn{{\bm{n}}}
\def\vo{{\bm{o}}}

\def\vt{{\bm{t}}}

\def\vx{{\bm{x}}}

\def\mI{{\bm{I}}}

\DeclareMathAlphabet{\mathsfit}{\encodingdefault}{\sfdefault}{m}{sl}
\SetMathAlphabet{\mathsfit}{bold}{\encodingdefault}{\sfdefault}{bx}{n}

\def\sR{{\mathbb{R}}}
\def\sS{{\mathbb{S}}}

\newcommand{\Ls}{\mathcal{L}}

\usepackage{xspace}
\makeatletter
\DeclareRobustCommand\onedot{\futurelet\@let@token\@onedot}
\def\@onedot{\ifx\@let@token.\else.\null\fi\xspace}

\def\eg{e.g\onedot}

\def\etal{et al\onedot}
\makeatother

\newcommand{\transpose}{^\mathsf{T}}

\newcommand{\dd}[2]{\frac{\text{d} #1}{\text{d} #2}}

\mainmatter
\title{SNeS: Learning Probably Symmetric\\Neural Surfaces from Incomplete Data}

\titlerunning{SNeS: Learning Probably Symmetric Neural Surfaces}
\author{Eldar Insafutdinov\thanks{Both authors contributed equally to this research.} \and
Dylan Campbell\printfnsymbol{1} \and
Jo\~ao F. Henriques \and
Andrea Vedaldi}
\authorrunning{E. Insafutdinov et al.}
\institute{University of Oxford, Oxford OX1 3PJ, UK
\email{\{eldar,dylan,joao,vedaldi\}@robots.ox.ac.uk}}

\begin{document}
\maketitle

\begin{abstract}
We present a method for the accurate 3D reconstruction of partly-symmetric objects.
We build on the strengths of recent advances in neural reconstruction and rendering such as Neural Radiance Fields (NeRF).
A major shortcoming of such approaches is that they fail to reconstruct any part of the object which is not clearly visible in the training image, which is often the case for in-the-wild images and videos.
When evidence is lacking, structural priors such as symmetry can be used to complete the missing information.
However, exploiting such priors in neural rendering is highly non-trivial:
while geometry and non-reflective materials may be symmetric, shadows and reflections from the ambient scene are not symmetric in general.
To address this, we apply a soft symmetry constraint to the 3D geometry and material properties, having factored appearance into lighting, albedo colour and reflectivity.
We evaluate our method on the recently introduced CO3D dataset, focusing on the car category due to the challenge of reconstructing highly-reflective materials.
We show that it can reconstruct unobserved regions with high fidelity and render high-quality novel view images.
\end{abstract}

\section{Introduction}%
\label{sec:introduction}

Photogrammetry has made substantial progress with recent advances in neural rendering \cite{tewari2021advances}.
Given a collection of posed images of an object, we can now use techniques such as COLMAP~\cite{schoenberger2016sfm} and NeRF~\cite{mildenhall2020nerf} to learn photo-realistic models of the object from which novel views can be generated.
Extensions such as NeuS~\cite{wang2021neus} and VolSDF~\cite{yariv2021volume} can also accurately recover the 3D shape of the object.
Many of these advances arise from using neural networks to represent the complex functions that describe the geometry and reflectance of the object.

\begin{figure}[!t]\centering
\begin{tabular}{cccc}
Seen (NeuS~\cite{wang2021neus}) & Seen (Ours) & Unseen (NeuS) & Unseen (Ours) \\
\includegraphics[width=29mm]{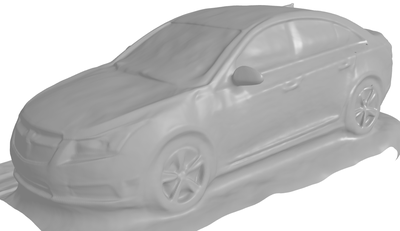} & \includegraphics[width=29mm]{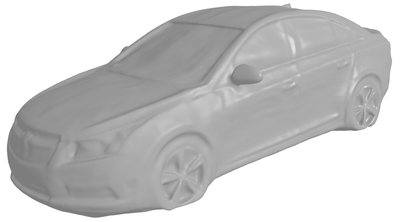} & \includegraphics[width=29mm]{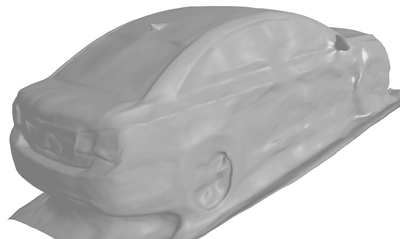} & \includegraphics[width=29mm]{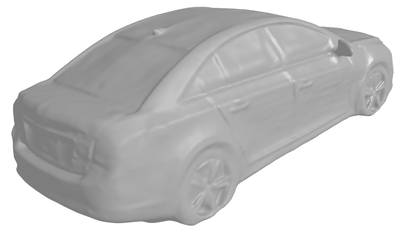}\\
\includegraphics[width=29mm]{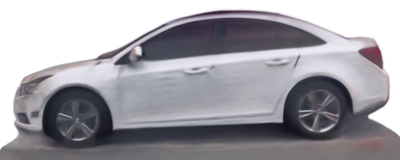} & \includegraphics[width=29mm]{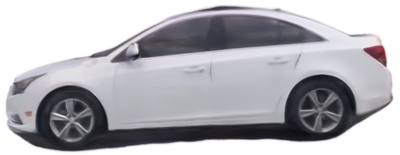} & \includegraphics[width=29mm]{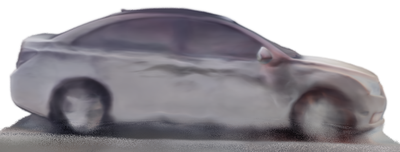} & \includegraphics[width=29mm]{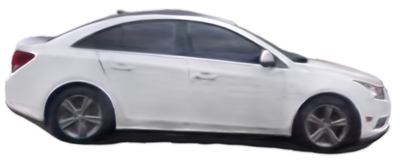}\\
\multicolumn{4}{c}{\includegraphics[width=121mm]{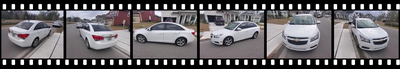}}
\end{tabular}
\caption{From a sequence of frames that view a car in passing, our Symmetric Neural Surfaces (SNeS) model simultaneously learns the parameters of a symmetry transformation from the data and applies the symmetry as a soft constraint to reconstruct the model, despite the significantly different view densities between the seen and unseen sides. The learned symmetry allows SNeS to share information across the model, resulting in more accurate reconstructions and higher-fidelity novel synthesised views.}%
\label{fig:splash}
\end{figure}

Despite such successes, significant practical limitations remain.
While networks often have excellent generalisation capabilities, in methods such as NeRF and NeUS they are overfitted to individual scenes (such as a \emph{single} 3D object).
As a result, such networks generalise poorly and are unable to predict the parts of the object that are not visible in the training images; instead, they require a large number of views capturing uniformly on all sides of the object.
This prevents applications in many realistic scenarios where only a limited and biased set of views is available, such as egocentric video or self-driving vehicles.

Bilateral symmetry is a strong geometric prior that applies approximately to many man-made and natural objects, and can be used to extrapolate beyond the field of view.
Unfortunately, symmetry is not directly applicable to current neural renderers, because they entangle potentially symmetric parts of the model (geometry, material) with ambient illumination and view-dependent effects (shadows, specularity, and reflections), which are not symmetric.
Our proposed approach, named \emph{Symmetric Neural Surfaces} (SNeS), thus decomposes a neural renderer's colour model into several components: material albedo (absorption), reflectivity, diffuse lighting, and reflected lighting.
These components are combined linearly (inspired by Phong shading~\cite{phong1975illumination}) and modelled with neural networks with different input constraints to ensure that they factorise correctly. For example, albedo only depends on the position and not on the viewpoint.
During training, we encourage symmetry for only a subset of these components, albedo and reflectivity, which are material-dependent.
We also apply symmetry to the geometry model, which is a neural surface model based on a signed distance function (SDF)~\cite{yariv2020multiview}.
Given the emphasis on bilateral symmetry and highly-reflective materials, our experiments are focused on vehicle reconstruction, which presents these unique challenges.
Our contributions are:
\begin{itemize}[nosep]
    \item an algorithm for reconstructing objects with arbitrary learned symmetries of a pre-defined type from incomplete observations;
    \item a technique for disentangling symmetric and asymmetric appearance; and
    \item a prior for handling violations of geometry and material symmetry.
\end{itemize}
We demonstrate high fidelity of reconstruction, both in visual appearance and in the accuracy of surface geometry, for parts of the objects that are unseen during training.
Our method achieves state-of-the-art results on the CO3D dataset \cite{reizenstein2021common}, improving the geometry estimates considerably compared to the baselines, especially on sequences where the view density between sides is unbalanced.

\section{Related Work}%
\label{sec:related_work}

The field of neural volume rendering has expanded rapidly in the last two years, with increasing photo-realism and reconstruction quality. We focus on the closest works, and refer readers to recent review papers for a complete account~\cite{tewari2020state,tewari2021advances}.

\paragraph{Neural volume rendering and reconstruction.}

Neural Radiance Fields (NeRF)~\cite{mildenhall2020nerf} and related approaches~\cite{lombardi2019neural,zhang2020nerf,barron2021mip,martin2021nerf,yu2021pixelnerf,wang2021ibrnet} generate images via a physic\-ally-based rendering process, where a ray is traced into the volume and neural network estimates of colour and density at sample points are integrated to render the pixel colour.
With careful network design or regularisation, such a model will be able to accurately reconstruct the scene's geometry as well as modelling view-dependent effects.
NeRF also introduced positional encoding, allowing MLPs to represent high frequency signals without increasing network capacity.
Our rendering pipeline is similar, but extended to model symmetries.

Many works investigate more sophisticated lighting models that reason about the transport and scattering of light through the volume, allowing relighting and material editing~\cite{bi2020neural,srinivasan2021nerv,boss2021nerd,chen2021dibr,zhang2021nerfactor,verbin2021refnerf}.
For example, NeRFactor~\cite{zhang2021nerfactor} converts a pre-trained NeRF model into a surface model, and optimises MLPs to represent light source visibility, surface normals, albedo, and the BRDF at any point on the surface, in addition to environment lighting, factoring appearance into material and lighting.
Ref-NeRF~\cite{verbin2021refnerf}, in contrast, trains a NeRF-like model from scratch, but replaces its parametrisation of outgoing radiance with one of reflected radiance to better model light transport, and estimates surface roughness to interpolate between blurry and sharp reflections.
Our model also decomposes appearance into material properties and lighting, using a Phong colour model~\cite{phong1975illumination} and a loss that encourages the diffusely-lit albedo of a surface point to match the ground truth on average, integrating over viewing directions.
Unlike existing work, this is motivated by symmetry learning, rather than editing applications, since lighting is typically asymmetric and impedes symmetry learning if ignored.

Many volume rendering approaches~\cite{mildenhall2020nerf,zhang2020nerf,barron2021mip} attempt to concentrate their samples near surfaces, \eg, by using stratified sampling.
Hybrid surface--volume representations~\cite{yariv2020multiview,oechsle2021unisurf,azinovic2021neural,wang2021neus,yariv2021volume} take this a step further by modelling surfaces directly, albeit implicitly, using occupancy~\cite{mescheder2019occupancy} or signed distance function (SDF)~\cite{park2019deepsdf} networks, combined with volume rendering for modelling view-dependent appearance.
This was motivated by the observation that NeRF, while able to handle sudden depth changes, is unable to learn high-fidelity surfaces from its implicit representation.
IDR~\cite{yariv2020multiview} represents the geometry as an SDF and uses a NeRF-like view-dependent head to estimate colour, which also receives the surface normal to better disentangle geometry and appearance.
However, the appearance network only receives one point per ray, at the first surface, which can cause the model to get stuck in local optima.
UNISURF~\cite{oechsle2021unisurf} relaxes this by using hierarchical sampling with root-finding in an occupancy field, allowing it to spread the gradient over multiple points, which nonetheless concentrate at the surface as training progresses.
A similar approach is taken by NeuS~\cite{wang2021neus} and VolSDF~\cite{yariv2021volume}, which represent surfaces as the zero-level set of an SDF and explore approaches for mapping signed distances to opacities.
Our work is a hybrid surface--volume approach of this type, since our aim is to reconstruct high-quality symmetric surfaces.
However, unlike previous work, we exploit additional structure in the data by learning symmetries and use them to share information between views.

\paragraph{Symmetry in 3D reconstruction.}
Symmetry cues have been used extensively in reconstruction, with shape-from-symmetry techniques enabling single-view reconstruction by using the reflected image as an additional view~\cite{gordon1990shape,Mukherjee94,Fawcett94,Huynh99,franccois2003mirror,thrun2005shape,sinha2012detecting,phillips2016seeing,chen2018autosweep}.
Symmetry detection has also been investigated~\cite{Forsyth92,sinha2012detecting}.
Of particular relevance to this work is the approach of Wu \etal~\cite{Wu20,Wu21,Wu21b}, who use reflective and rotational symmetries to recover shape, material properties and lighting from single images.
They enforce mirror symmetry by flipping internal representations of depth and albedo in image space, and estimate per-pixel probability of symmetry (confidence mask) to allow asymmetries.
Our work is inspired by this use of symmetry for reconstruction, and by the observation that asymmetric lighting must be removed to reason about appearance symmetries.
However, we target the task of multi-view reconstruction, apply a soft symmetry constraint in 3D directly (rather than in 2D), and learn the symmetry parameters to obviate the need for fronto-parallel images.

\section{Disentangled Neural Rendering}%
\label{sec:disentangling}

\begin{figure}[!t]\centering
    \resizebox{0.92\textwidth}{!}{\input{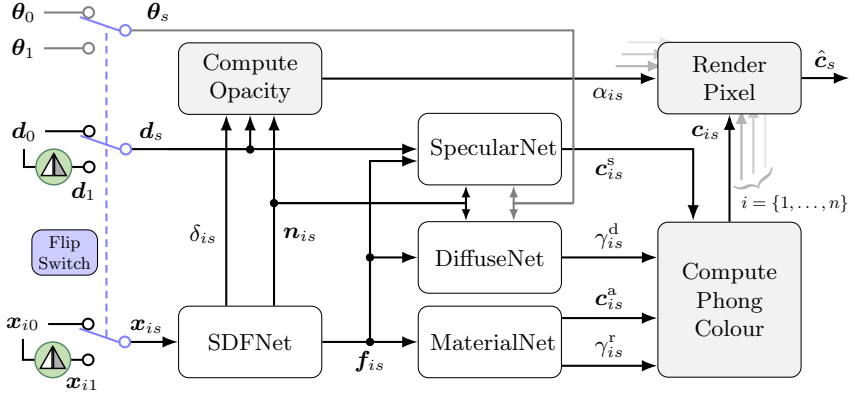}\unskip}
    \caption{The Symmetric Neural Surfaces (SNeS) model. For an input 3D point $\vx_{i}$ and direction vector $\vd$, the model estimates the geometry with an SDF network that generates a signed distance $\delta_{i}$, a normal vector $\vn_{i}$, and a feature vector $\vf_{i}$. The first two are used to compute the opacity $\alpha_{i}$ according to \cref{eqn:discrete_weight_fn}, which assigns high opacity to points near surfaces.
    The feature vector is passed to the appearance networks to compute the material properties of albedo colour $\vc_{i}^\text{a}$ and reflectivity $\gamma_{i}^\text{r}$, and the lighting properties of diffuse shading $\gamma_{i}^\text{d}$ and specular colour $\vc_{i}^\text{s}$. Lastly, the Phong model is used to compute the colour of the 3D point, and each sample along the ray is combined to render the pixel with colour $\hat{\vc}$. The subscript $s$ indicates whether the geometry, material and lighting components were computed with inputs that had undergone a symmetry transformation ($1$) or not ($0$), denoted by the triangular symbol. In each case, the lighting networks take different parameters $\theta$, since lighting is typically asymmetric.
    }%
    \label{fig:flowchart}
\end{figure}

In this section, we outline our disentangled neural rendering model that takes a collection of posed images and produces a signed distance function (SDF) of the geometry and an appearance model that can be queried from novel viewpoints.
In the subsequent section, we show how this baseline model can be used to learn symmetric neural surfaces.
A flowchart of our full model is shown in \cref{fig:flowchart}.

\subsection{Disentangling Geometry and Appearance}%
\label{sec:geometry_appearance}

Since the release of the NeRF model~\cite{mildenhall2020nerf}, there has been considerable research into improving the noisy surface reconstructions obtained by that method~\cite{yariv2020multiview,wang2021neus,oechsle2021unisurf}.
These have focused on replacing NeRF's density estimation network with a regularised SDF~\cite{yariv2020multiview,wang2021neus} or occupancy~\cite{oechsle2021unisurf} network.
We use NeuS~\cite{wang2021neus} as our baseline, since it effectively disentangles geometry and appearance, and is able to model fine structures.
For completeness, we recap the NeuS model now.

Given a set of images $\{ \mI_\ell \}$ with associated camera poses and intrinsic matrices, the task is to reconstruct the geometry and view-dependent appearance of the object or scene.
The geometry is represented implicitly as a signed distance function with zero-level set $\{\vx_i \in \sR^3 \mid \phi_\text{SDF} (\vx_i) = 0\}$ that coincides with opaque surfaces in the scene.
The map $\phi_\text{SDF} : \sR^3 \to \sR$, which converts a 3D point $\vx_i \in \sR^3$ to a signed distance $\delta_i$, is estimated with a fully-connected neural network.
The view-dependent appearance is also estimated by fully-connected neural network layers, parametrising the function $\phi_\text{colour} : \sR^3 \times \sS^3 \to \sR^3$, which maps a 3D point and view direction $\vd$ to a colour $\vc_i \in [0, 1]^3$.
Unlike NeuS, in this work the colour is estimated by a composition of functions to disentangle material and lighting properties, as shall be detailed in \cref{sec:material_lighting}.

To learn these functions from images, physically-based rendering accumulates colours along a pixel ray. The ray is parametrised as $\{\vx(t) = \vo + t\vd \mid t > 0 \}$ for a ray with camera centre $\vo$ and view direction $\vd$.
Rendering is performed by
\begin{align}
    \hat{\vc}(\vo, \vd) &= \int_{0}^{\infty} w(t) \, \vc(\vx(t), \vd)\,\mathrm{d}t,
    \label{eqn:render_continuous}
\end{align}
where $w$ is a weight function that satisfies $w(t) \geqslant 0$ and $\int_{0}^{\infty} w(t) \mathrm{d}t = 1$, and should be high near opaque surfaces.
In particular, $w$ should attain a local maximum at the zero-level set of the SDF, and should decay with distance from the camera.
NeuS derives an appropriate weight function with these properties,
\begin{align}
    w(t) &= \exp\left (-\int_0^t \rho(u)\,\mathrm{d}u \right) \rho(t) \text{, with} \;
    \rho(t) = \max \left \{ 0, \frac{-\dd{\sigma_\tau}{t} (\delta(t))}{\sigma_\tau (\delta(t))} \right \} ,
    \label{eqn:weight_fn}
\end{align}
where $\rho(t)$ is the opaque density function and $\sigma_\tau (x) = (1+ \exp (-\tau x))^{-1}$ is the sigmoid function parametrised by a learned scalar $\tau > 0$.
As can be seen, NeuS does not predict the volume density directly like NeRF, but rather computes the density using the predicted signed distances in closed form.
The learned scalar $\tau$ is proportional to the inverse standard deviation of the weight function (approximately a logistic density distribution), and controls the spread of the density about the zero-level crossing.
It adapts to the data during training, resulting in a more concentrated distribution over time.
This has two effects: colours of points near surfaces are assigned an increasingly high weight, and points are sampled increasingly close to surfaces, via an importance-sampling strategy.
We refer the reader to Wang~\etal~\cite{wang2021neus} for a detailed derivation.

A discrete approximation of the weight function follows from the quadrature technique used in NeRF~\cite{mildenhall2020nerf}.
For $n$ sampled points $\{\vx_i = \vo + t_i \vd \mid i = 1, \dots, n; \; t_i < t_{i+1} \}$ along the ray, their weights are given by
\begin{align}
    w_i &= \alpha_i \prod_{j=1}^{i-1} (1 - \alpha_j) \text{, with} \;
    \alpha_i = \max \left \{ 0, \frac{\sigma_\tau (\delta(t_i)) - \sigma_\tau (\delta(t_{i+1}))}{\sigma_\tau (\delta(t_i))} \right \},
    \label{eqn:discrete_weight_fn}
\end{align}
where the product term is the accumulated transmittance, and $\alpha_i$ is the discrete opacity. Note that to obtain the signed distance $\delta(t_i)$, the model uses the gradient (normal) vector to adjust the value of the nearest sampled signed distance.
The final colour is then rendered as
$
\hat{\vc} = \sum_{i=1}^{n} w_i \vc_i.
$

\subsection{Disentangling Material and Lighting Properties}%
\label{sec:material_lighting}

It is well-known that the NeRF colour formation model under-constrains the geometry, exhibiting a shape--radiance ambiguity where the training images can be perfectly explained by arbitrary geometry~\cite{zhang2020nerf}.
To impose a more realistic inductive bias on colour formation, without losing the flexibility and representation power of the unconstrained model, we disentangle the material and lighting properties using a Phong model~\cite{phong1975illumination}.
As we shall show, this is also a necessary requirement for learning symmetric geometries from the data.

We separate the apparent colour into material and lighting properties. Specifically, albedo colour and reflectivity (or inverse roughness) represent material, and diffuse shading (assuming a white diffuse illuminant) and specular colour represent lighting.
Here, we define the albedo as the average colour of a 3D point across viewpoints, under the scene lighting.
Our colour formation model is
\begin{align}
    \vc_{i} &= f_\text{Phong} \left( \gamma_{i}^\text{d}, \vc_{i}^\text{a}, \gamma_{i}^\text{r}, \vc_{i}^\text{s} \right) = \gamma_{i}^\text{d} (\vx_{i}, \vn_{i}) \, \vc_{i}^\text{a} (\vx_{i}) + \gamma_{i}^\text{r} (\vx_{i}) \, \vc_{i}^\text{s} (\vx_{i}, \vn_{i}, \vd_{i}),
    \label{eqn:phong}
\end{align}
where $\vc_{i} \in [0, 1]^3$ is the estimated colour of the 3D point $\vx_{i}$, $\gamma_{i}^\text{d} \in [0, 2]$ is the diffuse lighting coefficient, $\vc_{i}^\text{a} \in [0, 1]^3$ is the lighting-invariant albedo colour of the material, $\gamma_{i}^\text{r} \in [0, 1]$ is the material reflectivity, and $\vc_{i}^\text{s} \in [0, 1]^3$ is the specular colour of the reflected light.
We see that the material properties depend on the geometry only, while the lighting depends additionally on the normal vector (diffuse lighting with self-shadows) and the viewing direction (specular colour).
A more constrained parametrisation would learn the specular colour from the viewing ray reflected about the surface normal~\cite{verbin2021refnerf}.
However, we found that this significantly over-smoothed the SDF model.
While this colour model has the capacity to disentangle material and lighting properties, it needs to be regularised in order to do so.
However, some objects, such as cars, are highly specular, making it undesirable to regularise the reflectivity.
We instead encourage the diffusely-lit colour $\gamma_{i}^\text{d} \vc_{i}^\text{a}$, rendered along the ray, to match the ground-truth colour, as we shall detail in the next section.
This acts to average the colour of a surface location across all viewing directions.
In practice, the appearance networks also depend on a feature vector from the SDF network, encoding the geometric context of the 3D point~\cite{yariv2020multiview}.

\section{Symmetric Neural Surfaces}%
\label{sec:symmetry}

\begin{figure}[!t]\centering
    \begin{minipage}{0.5\linewidth}
       \begin{subfigure}[]{\textwidth}\centering
	   \includegraphics[width=\textwidth]{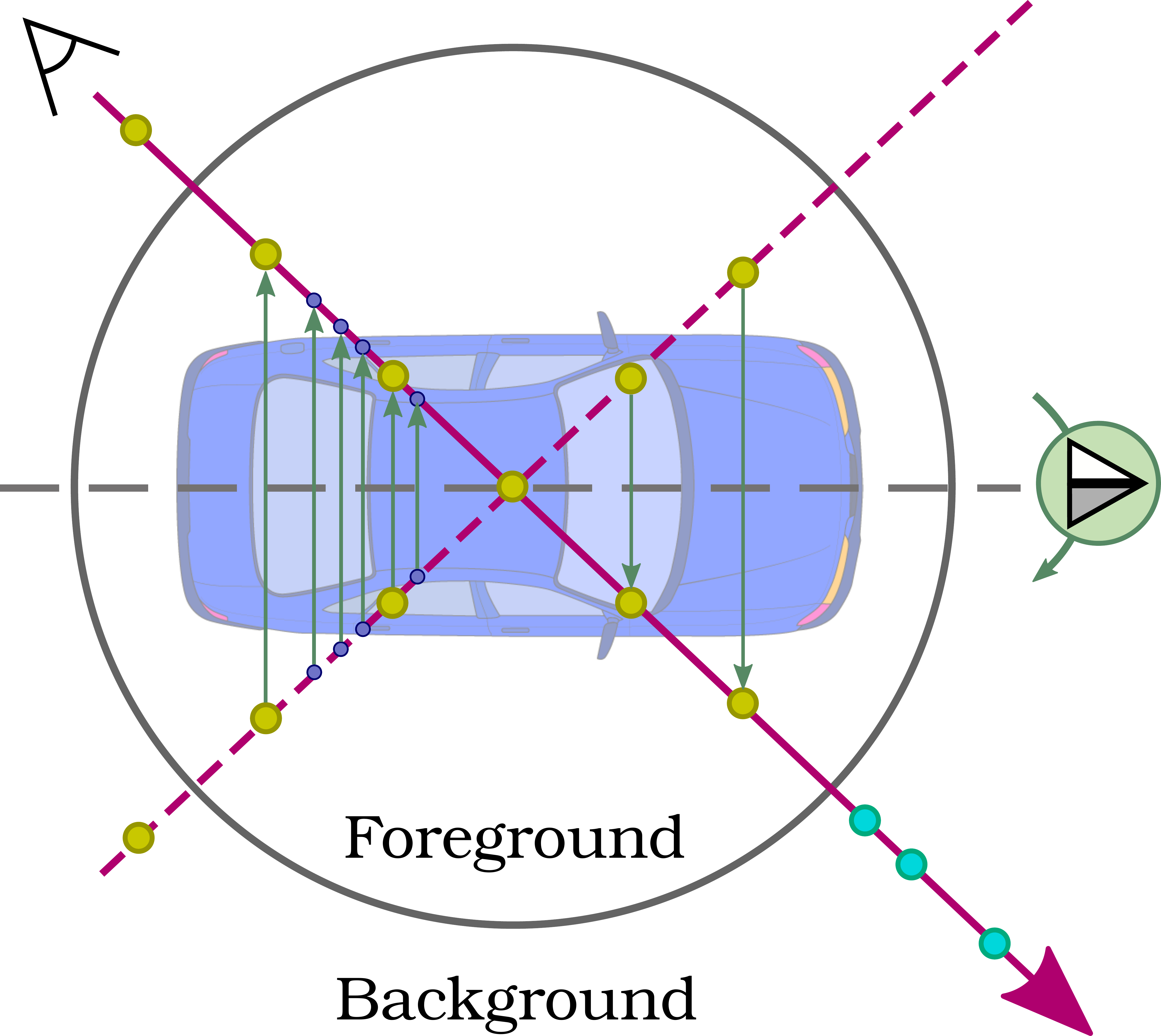}
		\caption{Symmetry transformation}%
		\label{fig:schematic}
        \end{subfigure}
	\end{minipage}\hfill
	\begin{minipage}{0.5\linewidth}\centering
	\begin{subfigure}[]{\textwidth}\centering
		\includegraphics[width=0.8\textwidth]{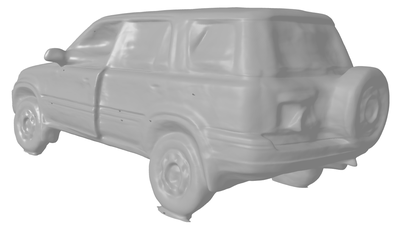}
		\caption{3D reconstruction with asymmetry}%
		\label{fig:asymmetry_1}
	\end{subfigure}\vfill
	\begin{subfigure}[]{0.8\textwidth}\centering
		\includegraphics[width=0.70\textwidth]{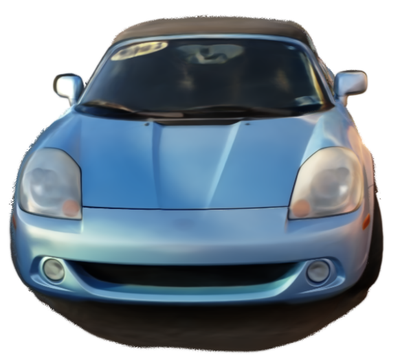}
		\caption{Novel view with asymmetry}%
		\label{fig:asymmetry_2}
	\end{subfigure}
	\end{minipage}
    \caption{
    \subref{fig:schematic}~Applying a symmetry transformation for physically-based rendering.
    The SNeS algorithm scales the object of interest to fit inside the unit sphere, where it is modelled by an SDF network with appearance heads, while the region outside the sphere is represented by a NeRF++ background model~\cite{zhang2020nerf}. Here, yellow dots denote points sampled coarsely along the ray, small blue dots denote points importance-sampled near dominant surfaces, green dots denote points inversely-sampled in the background, and the horizontal dashed line denotes the plane of (reflection) symmetry. The symmetry induces a transformation on the point samples inside the sphere, and the transformed points are used to compute the geometry and material properties. These components are combined with the diffuse and specular lighting estimates from the source ray to form a colour estimate. If the symmetry holds, and is accurately estimated, the resulting colour should match the source colour.
    \subref{fig:asymmetry_1}~SNeS reconstruction showing that geometry asymmetries (spare tire, slightly-open door) are conserved.
    \subref{fig:asymmetry_2}~SNeS novel view showing that appearance asymmetries (windshield sticker, lighting) are conserved.
    }\label{fig:birdseye}
\end{figure}

The model described thus far is unable to take advantage of known or suspected symmetries. We define a symmetry as an arbitrary coordinate transformation, especially an affine transformation such as a reflection, rotation, translation, or scaling, that confers an invariance.
To share information across symmetries, we explicitly model and optimise the transformation parameters and use the map induced by the symmetry to aggregate information in 3D.
However, not all information is symmetric.
Cars, for example, tend to have a bilateral symmetry in their geometry and material properties, but the lighting of the scene is rarely symmetric.
There are also exceptions that break the geometric and material symmetry, such as asymmetrically-positioned spare tyres (geometric) and stickers (material colour), as shown in \cref{fig:birdseye} (c).
Due to these real-world partial symmetries, it is best implemented as a soft constraint mediated by loss functions, rather than a hard constraint.

Our framework has the flexibility to handle multiple arbitrary and localised symmetries. Spatially-restricted symmetries can be useful for modelling parts of an object or scene that are locally-symmetric, like the wheels of a car.
In this work we focus on the common case of reflection (bilateral) symmetry, and localise the predicted symmetry to the unit sphere, to avoid symmetrising the background. A diagram of our approach is shown in \cref{fig:birdseye}.
In the following, we denote the original ray, and everything computed with respect to it, as ``source'', and the symmetry-transformed ray as ``transformed''.

\subsection{Parametrising Symmetry}%
\label{sec:symmetry_parametrisation}

We parametrise a symmetry as a coordinate transformation with form
$
T_\text{c}^{-1} S T_\text{c}$, where $T_\text{c} =
\big(\begin{smallmatrix}
    R_\text{c} & \vt_\text{c} \\
    \mathbf{0} & 1
\end{smallmatrix}\big)
$
is the learned rigid transformation matrix from world coordinates to the canonical coordinates of the symmetry, defining its plane or axis, and $S$ is the symmetry transformation in canonical coordinates.
In this work, we consider a single bilateral reflection symmetry about the $XZ$ plane in canonical coordinates, and so the symmetry transformation matrix is given by $S = I - 2 e_2 e_2\transpose$, where $e_i$ is the $i\textsuperscript{th}$ 4D unit basis vector.

We apply the transformation to the source points $\vx_0^\text{h}$ in homogeneous coordinates. The direction vectors $\vd_0^\text{h}$ also undergo a symmetry transformation, although they are translation-invariant so are not affected by changes in position. This is implemented by the homogeneous coordinates, since directions are points at infinity with final coordinate equal to 0. Thus we obtain
\begin{align}
    \vx_{i1}^\text{h} &= T_\text{c}^{-1} S T_\text{c} \vx_{i0}^\text{h} \label{eqn:symmetry_point}\\
    \vd_{i1}^\text{h} &= T_\text{c}^{-1} S T_\text{c} \vd_{i0}^\text{h}. \label{eqn:symmetry_direction}
\end{align}

\subsection{Learning Symmetric Geometry and Material}%
\label{sec:symmetry_learning}

To encourage symmetric points to have the same geometry and material properties, we compute these quantities at both the source and transformed points, and compose them with predictions from the corresponding lighting model.
Thus, for each point, we obtain the source colour
$
\vc_{i00} = \gamma_{i0}^\text{d} \vc_{i0}^\text{a} + \gamma_{i0}^\text{r} \vc_{i0}^\text{s}
$
and the symmetry-transformed colour
$
\vc_{i11} = \gamma_{i1}^\text{d} \vc_{i1}^\text{a} + \gamma_{i1}^\text{r} \vc_{i1}^\text{s}
$.
The lighting models for the source and symmetry-transformed paths do not share weights, since lighting is rarely symmetric.
The resulting point colours are rendered along the ray, and compared to the ground-truth source pixel colour.
If symmetry is valid at that pixel, and is accurately estimated, the error should be low.
However, most objects and scenes are not perfectly symmetric, and so symmetry should not be enforced when better visual evidence is available.
Therefore, we penalise the error of the symmetry-transformed colour at a discount compared to the source colour.

We also mix the source and the transformed components, generating hybrid colours.
This acts to supervise the transformed lighting network to emulate the source lighting network, up to the symmetry transformation.
Without these terms, the lighting networks may diverge, allowing the network to explain away deviations from symmetry as fake perturbations in lighting.
Specifically, we form the hybrid point-wise colours $\vc_{i01} = \gamma_{i1}^\text{d} \vc_{i0}^\text{a} + \gamma_{i0}^\text{r} \vc_{i1}^\text{s}$ and $\vc_{i10} = \gamma_{i0}^\text{d} \vc_{i1}^\text{a} + \gamma_{i1}^\text{r} \vc_{i0}^\text{s}$, render these along the ray, and compute the colour error as before.

It is important to disentangle the material and lighting, since the former is usually asymmetric.
This means that simply applying symmetry to the NeRF colour model would not work, since the colour is entangled with a systematic nuisance variable.
Another strategy to help estimate the symmetry parameters is to learn the ground plane simultaneously and enforce orthogonality between the ground plane and the symmetry plane.
To do so, we model the foreground as a joint SDF, which consists of the minimum of the object's SDF and a ground plane SDF (an infinite plane).
This allows the SDF network to spend more capacity on the object, and enables ground removal without post-processing.

\subsection{Loss functions}%
\label{sec:losses}

To fit our model, we minimise the error between the rendered and ground-truth pixels while regularising the SDF\@.
No 3D supervision is used, beyond the known camera poses.
We optimise the network parameters, symmetry transformation parameters, and the scalar $\tau$ that controls the variance of the density near surfaces.
The per-pixel colour loss is given by
\begin{align}
    \Ls_{jk}^\text{colour} &= \tfrac{1}{3} \| \hat{\vc}_{jk} - \vc \|_1,
    \label{eqn:loss_colour}
\end{align}
where $\vc$ is the ground-truth colour and $\hat{\vc}_{jk}$ is the predicted colour.
The indices $jk$ indicate whether the colour prediction uses the source or symmetry-transformed geometry and material properties ($j$), and lighting ($k$).
This is the mechanism by which symmetry is encouraged in regions with visual evidence to the contrary.

We also use two additional losses with the same form as \cref{eqn:loss_colour}.
The first is a diffuse colour loss $\Ls_{jk}^\text{diffuse}$ where the predicted colour is rendered without the specular components, that is, the pixel colour is rendered from point colours $\vc_{i}^\text{diffuse} = \gamma_{i}^\text{d} \vc_{i}^\text{a}$.
This encourages the network to disentangle the diffuse and specular components, setting the diffuse colour of a given surface location to the average colour across all viewing directions.
This is important for symmetry, since the specular colour is usually not symmetric, so assisting the network to disentangle it can speed up convergence.
The second is a symmetric lighting loss $\Ls_{jk}^\text{lighting}$ that applies a weak prior to the model to prefer symmetric lighting in the absence of contrary evidence.
It applies the same colour loss as \cref{eqn:loss_colour}, but with the source lighting networks receiving symmetry-transformed inputs.
This acts to apply symmetric lighting, which is generally incorrect, except at midday.
Nonetheless, in the absence of image evidence, this prior provides a more naturalistic appearance.
However, this loss should not be applied for quantitative analysis of unseen sides, because applying the symmetric lighting model is likely to be more detrimental than applying a baseline lighting model.
For example, it may apply direct sunlight and specular reflections on the shadowed side of the object, which may look qualitatively convincing, but will be quantitatively poor.

Finally, we regularise the SDF network by applying an Eikonal loss~\cite{gropp2020implicit} at the $n$ sampled points along the ray, which encourages a unit gradient SDF:
\begin{align}
    \Ls_{j}^\text{eikonal} &= \frac{1}{n} \sum_i^{n} (\|\nabla \phi_\text{SDF} (\vx_{ij})\|_2 - 1)^2.
    \label{eqn:loss_eikonal}
\end{align}
The total per-pixel loss is given by
\begin{align}
    \Ls &= \sum_{j,k} (1 + (\lambdasym - 1)j) \left( \Ls_{jk}^\text{colour} + \lambda^\text{d} \Ls_{jk}^\text{diffuse} + \lambda^\text{l} \Ls_{jk}^\text{lighting} + \lambda^\text{e} \Ls_{j}^\text{eikonal} \right),
    \label{eqn:loss_total}
\end{align}
where $\lambdasym \in [0, 1]$ is the symmetricity factor that determines a prior on how symmetric an object or scene is expected to be, and the other $\lambda$ factors denote the weights assigned to the remaining losses.

\section{Results}%
\label{sec:results}

\subsection{Experimental Setup}%
\label{sec:setup}

\paragraph{Dataset.}
We evaluate our method on the cars subset of the recent Common Objects in 3D (CO3D) dataset~\cite{reizenstein2021common}, released under the BSD License.
CO3D is a large-scale multi-view image dataset with ground-truth camera pose, intrinsics, depth map, object mask, and 3D point cloud annotations, collected in-the-wild by outdoor video capture.
This real-world dataset is particularly challenging for reconstruction algorithms, having highly reflective (non-Lambertian) and low-textured surfaces, such as mirrors, dark windows, and metallic paint.
Moreover, only 64\% of the test sequences circumnavigate the object, with many seeing only one side of the car.
This incomplete data motivates the use of symmetry for completing the reconstruction of partially-symmetric objects.
Additional nuisance factors include significant motion blur from the handheld cameras, auto-exposure,
and adverse weather, including fog and rain.
One of the consequences of this challenging data is that the ground-truth point clouds and depth maps are sparse, very noisy, and contain many outliers, %
and 8\% of test object masks entirely miss the object.
This makes evaluating the reconstructed geometry, especially fine details, quite difficult.
For the task of single-scene 3D reconstruction and novel view synthesis, the `car' category has 22 test scenes with 102 frames each.
We present results on other partly-symmetric categories in the appendix.

\paragraph{Metrics.}
We report five metrics to measure visual and geometric quality: the peak signal-to-noise ratio (PSNR), the mean squared colour error (MSE), and the perceptual LPIPS distance~\cite{zhang2018unreasonable} between the masked predicted and ground-truth novel-view images; the mean absolute error (MAE) between the masked predicted and ground-truth depth maps; and the intersection-over-union (IoU) between the predicted and ground-truth object masks.

\paragraph{Baselines.}
We compare with two state-of-the-art baselines for novel-view synthesis and 3D reconstruction in unbounded, real-world scenes: NeRF++~\cite{zhang2020nerf} and NeuS~\cite{wang2021neus}.
We do not compare with the state-of-the-art classical multi-view stereo algorithm COLMAP~\cite{schoenberger2016sfm}, because the dataset's ground-truth point clouds and depth maps were obtained using this algorithm and are extremely noisy and sparse for this reflective and low-texture category.
We focus on two strong and well-regarded baselines to avoid the evaluation becoming prohibitively expensive (each baseline trains for at least 24h on a single GPU).

\paragraph{Implementation details.}
Following prior art~\cite{yariv2020multiview,wang2021neus}, we implement the SDF network $\phi_\text{SDF}$ as an 8-layer MLP with hidden dimension 256, position-encoded inputs (6 frequencies)~\cite{mildenhall2020nerf}, and geometric initialisation for the network weights~\cite{atzmon2020sal}. %
The material, diffuse, and specular networks are also implemented as MLPs with 4/2/4 hidden layers, with a 4-frequency positional encoding on the normal and view directions.
NeRF++~\cite{zhang2020nerf} is used as the background model.
We follow the hierarchical sampling strategy of NeuS~\cite{wang2021neus} with 64 coarse, 64 fine, and 32 background samples per ray, with 1024 rays sampled per batch.
We optimise the network with Adam~\cite{kingma2014adam} and an initial learning rate of 5e-4 and train for 300K iterations on a single GPU\@.
Unless otherwise stated, we use the hyperparameters $[\lambdasym, \lambda^\text{d}, \lambda^\text{l}, \lambda^\text{e}] = [0.1, 0.01, 0.001, 0.1]$.
Complete implementation details are reported in the supplementary material, and we will release the code publicly.

\subsection{Random Test Split}%
\label{sec:random}

\begin{table}[!t]\centering
\caption{Results on the random and structured test splits of the CO3D cars dataset~\cite{reizenstein2021common}.
We report the peak signal-to-noise ratio (PSNR), mean squared error (MSE), and LPIPS distance between the estimated and ground-truth masked images, the mean absolute error (MAE) between the estimated and ground-truth masked depth maps, and the intersection-over-union (IoU) of the estimated and ground-truth masks.
}\label{tab:co3d}
\setlength{\tabcolsep}{0pt} %
{\scriptsize
\begin{tabularx}{\linewidth}{@{}l C C C C C | C C C C C@{}}
\toprule
& \multicolumn{5}{c|}{Random Split (overlapping viewpoints)} & \multicolumn{5}{c}{Structured Split (biased viewpoints)}\\
& PSNR & MSE & LPIPS & MAE & IoU & PSNR & MSE & LPIPS & MAE & IoU\\
& RGB & RGB & RGB & Depth & Mask & RGB & RGB & RGB & Depth & Mask\\
Method & $\uparrow$ & $\downarrow$ & $\downarrow$ & $\downarrow$ & $\uparrow$ & $\uparrow$ & $\downarrow$ & $\downarrow$ & $\downarrow$ & $\uparrow$\\
\midrule
NeRF++ \!\!\cite{zhang2020nerf} & 21.4 & 0.007 & 0.407 & 0.222 & -- & 13.9 & 0.041 & 0.581 & 0.177 & --\\
NeuS \cite{wang2021neus} & \textbf{23.3} & \textbf{0.005} & 0.355 & 0.108 & 0.523 & 13.4 & 0.046 & 0.556 & 0.105 & 0.566\\
SNeS (ours) & \textbf{23.3} & \textbf{0.005} & \textbf{0.348} & \textbf{0.086} & \textbf{0.787} & \textbf{14.1} & \textbf{0.039} & \textbf{0.503} & \textbf{0.077} & \textbf{0.906}\\
\bottomrule
\end{tabularx}%
}
\end{table}

For this experiment, we use the train--test split provided by the dataset for single scene experiments (``test\_known'' and ``test\_unseen'')~\cite{reizenstein2021common}, assigned at random from the frames of the video. This evaluates the model's ability to interpolate between a dense set of views.
This is the standard mode for evaluating novel view synthesis algorithms.
Note that only 64\% (14) of the video sequences entirely encircle the object of interest, with the remainder having coverage of as little as 135$^\circ$.
Our method is able to reconstruct the unseen sides, though we are unable to evaluate this as the requisite ground-truth is not present in the dataset.
The results are shown in \cref{tab:co3d}, and indicate that applying symmetry does not harm the performance of the baseline model, and indeed improves the geometry.
This suggests that the model is able to learn the symmetry and integrate information from both sides of the object to improve the geometry estimate.

\subsection{Structured Test Split}%
\label{sec:structured}

\begin{figure}[!t]\centering
	\begin{subfigure}[]{0.25\textwidth}\centering
	   \includegraphics[width=\textwidth]{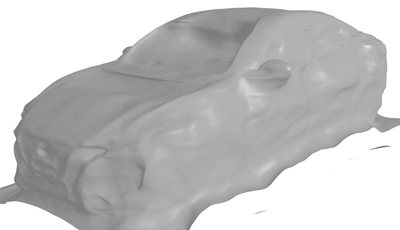}
		\label{fig:ablation_1}
	\end{subfigure}\hfill
	\begin{subfigure}[]{0.25\textwidth}\centering
		\includegraphics[width=\textwidth]{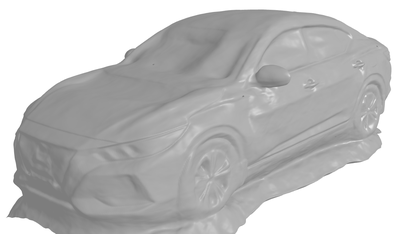}
		\label{fig:ablation_2}
	\end{subfigure}\hfill
	\begin{subfigure}[]{0.25\textwidth}\centering
		\includegraphics[width=\textwidth]{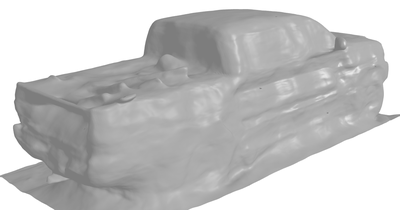}
		\label{fig:ablation_3}
	\end{subfigure}\hfill
	\begin{subfigure}[]{0.25\textwidth}\centering
		\includegraphics[width=\textwidth]{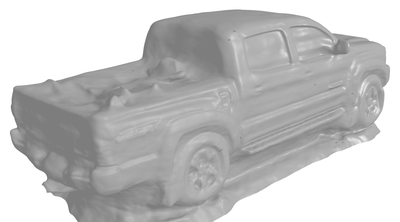}
		\label{fig:ablation_4}
	\end{subfigure}\vfill
	
	\begin{subfigure}[]{0.25\textwidth}\centering
	   \includegraphics[width=\textwidth]{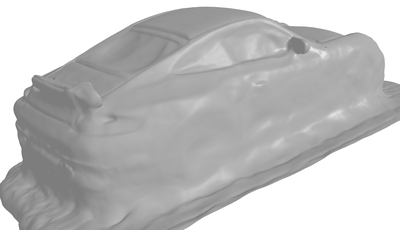}
		\caption{NeuS}%
		\label{fig:ablation_1}
	\end{subfigure}\hfill
	\begin{subfigure}[]{0.25\textwidth}\centering
		\includegraphics[width=\textwidth]{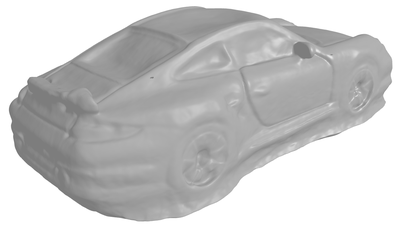}
		\caption{SNeS}%
		\label{fig:ablation_2}
	\end{subfigure}\hfill
	\begin{subfigure}[]{0.25\textwidth}\centering
		\includegraphics[width=\textwidth]{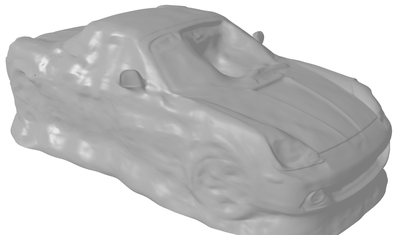}
		\caption{NeuS}%
		\label{fig:ablation_3}
	\end{subfigure}\hfill
	\begin{subfigure}[]{0.25\textwidth}\centering
		\includegraphics[width=\textwidth]{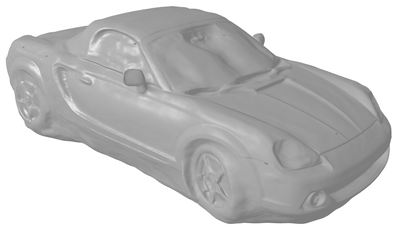}
		\caption{SNeS}%
		\label{fig:ablation_4}
	\end{subfigure}%
    \caption{Qualitative results on the structured test split of the CO3D cars dataset~\cite{reizenstein2021common}.}%
    \label{fig:structured_qualitative}
\end{figure}

\begin{figure}[!t]\centering

    \begin{tabular}{ccccc}
    \rotatebox[origin=l]{90}{\scriptsize \hspace{7pt} NeuS} &
    \includegraphics[height=14.5mm]{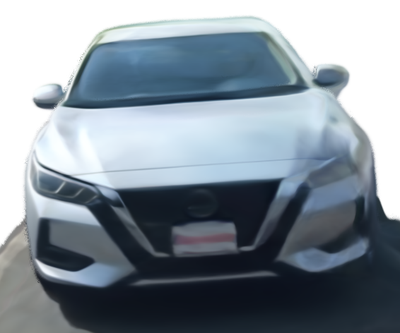} & \includegraphics[height=14.5mm]{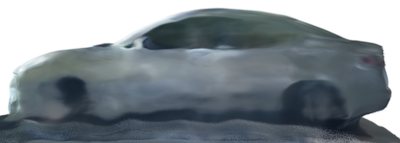} & \includegraphics[height=14.5mm]{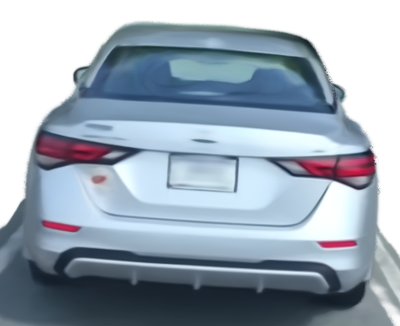} & \includegraphics[height=14.5mm]{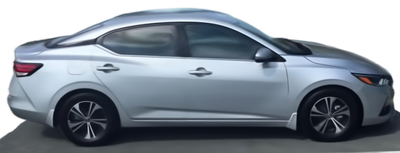}\\
    
    \rotatebox[origin=l]{90}{\scriptsize \hspace{7pt} SNeS} &
    \includegraphics[height=14.5mm]{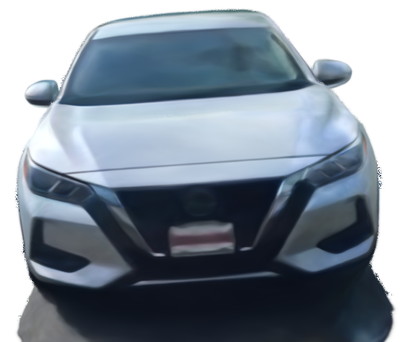} & \includegraphics[height=14.5mm]{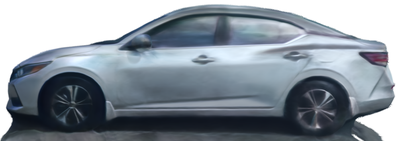} & \includegraphics[height=14.5mm]{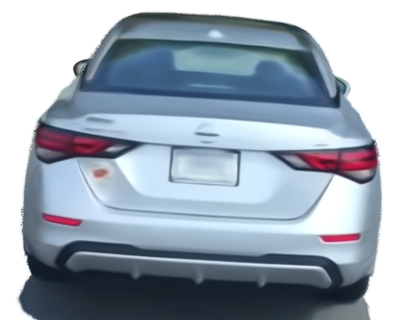} & \includegraphics[height=14.5mm]{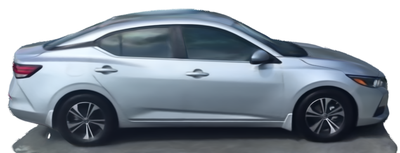}\\
    
    \rotatebox[origin=l]{90}{\scriptsize SNeS-Albedo} &
    \includegraphics[height=14.5mm]{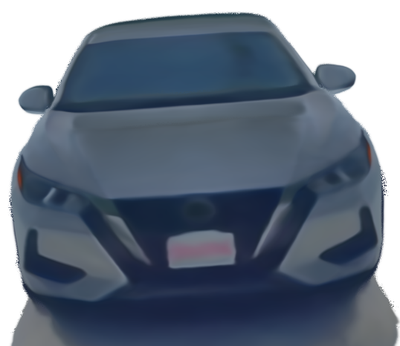} & \includegraphics[height=14.5mm]{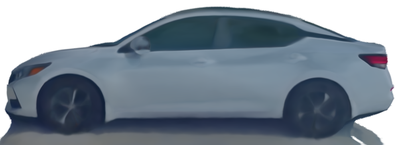} & \includegraphics[height=14.5mm]{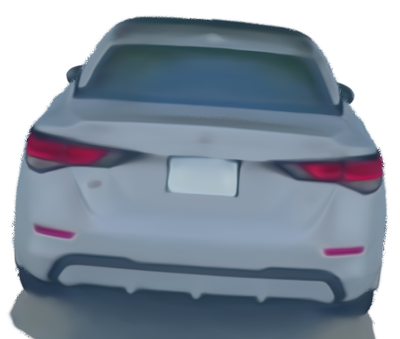} & \includegraphics[height=14.5mm]{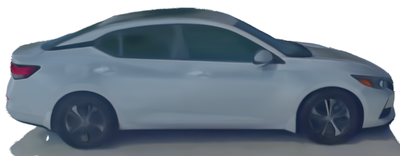}\\
    \end{tabular}

    \caption{Novel view renderings of the partly-observed (left) and fully-observed (right) sides. Top row: NeuS. Middle row: SNeS. Bottom row: SNeS albedo maps.}%
    \label{fig:novel_views}
\end{figure}

We propose a new train--test split that simulates the common situation where one side of an object is observed more thoroughly than the other.
This tests the model's ability to handle variable view densities and incomplete information.
To do so, we select the 14 test scenes where the camera circumnavigates the object, and define a test split that sets aside all camera poses within a 130$^\circ$ sector emanating from the object's centre, approximately perpendicular to the plane of bilateral symmetry.
Thus, one side of the car is only seen obliquely.
From the set aside poses, we systematically sample 8 test frames.
This setting makes it possible for existing methods to reconstruct both sides of the object, but tests how well they are able to reconstruct the side that is viewed less fully.
The results are shown in \cref{tab:co3d}.
Our method consistently outperforms the NeuS baseline on the novel view synthesis metrics and significantly improves the depth accuracy on the unseen side, validating the effectiveness of our approach.
Qualitative comparisons are shown in \cref{fig:structured_qualitative,fig:novel_views}, demonstrating high-fidelity reconstructions and synthesised views on the unseen side.
We include additional high-resolution qualitative results in the supplementary material, including a comparison of the different appearance components (material and lighting).

\subsection{Ablation Study}%
\label{sec:ablation}

\begin{table}[!t]\centering
\caption{Ablation study on a random subset of our structured test split of the CO3D cars dataset~\cite{reizenstein2021common}.
We report the peak signal-to-noise ratio (PSNR), mean squared error (MSE), and LPIPS distance between the estimated and ground-truth masked images, the mean absolute error (MAE) between the estimated and ground-truth masked depth maps, and the intersection-over-union (IoU) of the estimated and ground-truth masks.
}%
\label{tab:ablation}
\setlength{\tabcolsep}{0pt} %
{\scriptsize
\begin{tabularx}{\linewidth}{@{}l C C C C C@{}}
\toprule
Method & PSNR RGB $\uparrow$ & MSE RGB $\downarrow$ & LPIPS RGB $\downarrow$ & MAE Depth $\downarrow$ & IoU Mask $\uparrow$\\
\midrule
SNeS (ours) & \textbf{14.3} & \textbf{0.0372} & 0.564 & 0.0706 & 0.894\\
$+\Ls^\text{lighting}$ & 13.7 & 0.0425 & 0.585 & 0.0685 & 0.914\\
$-\Ls^\text{diffuse}$ & \textbf{14.3} & \textbf{0.0372} & 0.566 & 0.0722 & \textbf{0.917}\\
$-\Ls^\text{col}$ & 13.7 & 0.0422 & 0.576 & 0.0782 & 0.906\\
\bottomrule
\end{tabularx}%
}
\end{table}

\begin{figure}[!t]\centering
    \setlength{\tabcolsep}{0pt}
    \begin{tabular}{ccccc}
    \includegraphics[width=0.2\linewidth]{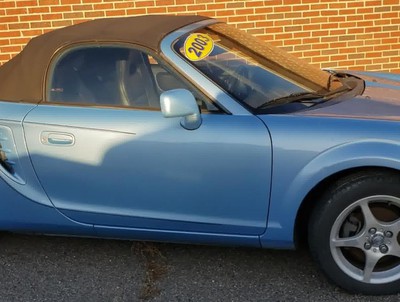} &
    \includegraphics[width=0.2\linewidth]{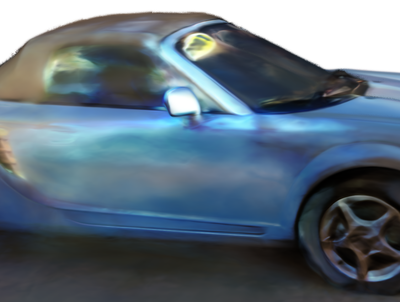} &
    \includegraphics[width=0.2\linewidth]{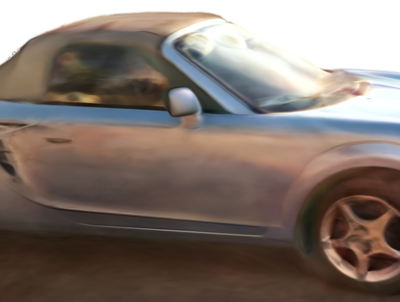} &
    \includegraphics[width=0.2\linewidth]{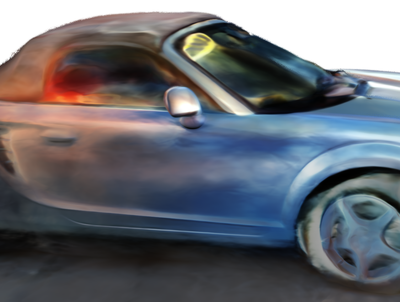} &
    \includegraphics[width=0.2\linewidth]{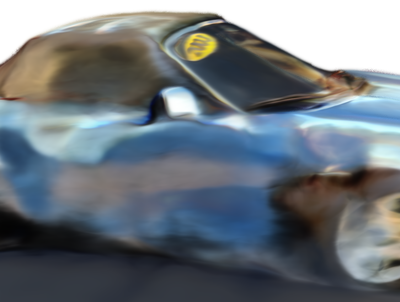}\\[-0.3em]
    \scriptsize Ground-truth &
    \scriptsize Ours &
    \scriptsize $+\Ls^\text{lighting}$ &
    \scriptsize $-\Ls^\text{diffuse}$ &
    \scriptsize $-\Ls^\text{col}$
    \end{tabular}
    \caption{Qualitative ablation study. Novel view renderings of the unseen side.}
    \label{fig:ablation}
\end{figure}

To investigate the effect of the different components of our model, we ablate its performance on a subset of 4 randomly selected scenes from our structured test split of the CO3D cars dataset, as shown in \cref{tab:ablation} and \cref{fig:ablation}.
We ablate with respect to the model without the lighting loss (ours), since this loss is designed to produce qualitatively convincing renders in the absence of image evidence, but is unlikely to be quantitatively (visually) accurate in those areas.
We indeed see that the symmetric lighting loss has a detrimental effect on the image-based results, predominantly in situations where direct sunlight is applied to the shadowed side of the car and vice versa.
However, the resulting renders are qualitatively preferable, as shown at high resolution in the appendix.
We verify that removing the diffuse colour loss harms the geometry, since it helps decouple the symmetric and asymmetric properties facilitating symmetry learning.
Finally, we show that removing the symmetry loss $\Ls^\text{col}$ (including mixed terms) significantly reduces the visual and geometric quality.

\section{Discussion and Limitations}%
\label{sec:limitations}

One limitation of the approach is that it is only beneficial for objects or scenes with significant symmetries.
However, this is not as restrictive as it might seem.
While the natural world rarely has large-scale symmetries, they abound in the human environment, in architecture and object design.
For example, out of the CO3D dataset, 90\% of the categories have at least one major symmetry, such as ball, baseball bat, bench, bicycle, book, bottle, and bowl.
More significant limitations of the approach, then, are that the type and number of symmetries must be specified in advance, that the symmetry has to be significant enough to be learnable from the data, and that the initialisation of the symmetry plane or axis must be good enough to avoid the network getting trapped in a local optimum.
An alternative approach, such as multiple initialisations, may be necessary to prevent the latter in some cases.
Another limitation of the approach is that it requires a significant number of views, even with the reductions facilitated by the symmetry.
This is because it can be difficult to optimise the symmetry parameters, such as finding the reflection plane, without reasonable view coverage.
This could be mitigated by learning about symmetries from a collection of scenes, such that a single view may be enough to partially constrain the symmetry plane parameters \cite{zhou2021nerd}.
Our approach also relies on good camera estimates.
While this requirement can be relaxed~\cite{wang2021nerfmm}, additional unknown variables are likely to make the symmetry parameters more difficult to estimate.
Finally, our approach does not explicitly handle symmetries that are present at different scales or resolutions.
For example, a decorated cake or a pizza is symmetric at one scale, but may violate that symmetry when considering the finer details.
Applying our framework to a multi-scale reconstruction method like Mip-NeRF~\cite{barron2021mip} would allow us to apply symmetry differently at different resolutions.

\section{Conclusion}%
\label{sec:conclusion}

We have presented a 3D reconstruction and novel-view synthesis method for partly-symmetric objects, which learns symmetry parameters from a collection of posed images and uses the learned symmetry to share information across the model.
This reduces the need for dense multi-view coverage of the object, making it suitable for use on in-the-wild data like the CO3D dataset. We demonstrated our algorithm on objects that exhibit bilateral symmetry at most locations---cars---and show that it can reconstruct unobserved regions with high fidelity.

\end{document}


\maketitle
\appendix

\section{Characterisation of Symmetries}%
\label{sec:symmetry}

While in the main paper we represent symmetry as a general coordinate transformation, or rather as a symmetry transformation in a canonical coordinate frame, that is,
$T_\text{c}^{-1} S T_\text{c}$,
here we provide a taxonomy of some specific symmetry types and how they can be implemented.
Some terminology and formulae are adapted from Rosen~\cite{rosen1975symmetry} and Thrun \& Wegbreit~\cite{thrun2005shape}.
Note that here we only consider point transformations; direction vectors are invariant to translation and so only undergo the rotation or scaling parts of the transformation.
\begin{enumerate}
    \item \textbf{Reflection symmetry:} whereby a point is mapped to another point (a bijection) on the opposite side of a plane, line or point.
    \begin{enumerate}
        \item Planar reflection: for a plane defined by a normal vector $\vn$ and scalar distance from the origin $\Delta$, a point $\vx$ is mapped to
        \begin{align}
            \vx' &= \vx - 2\vn (\vn\transpose \vx - \Delta)\\
            &= (I - 2\vn \vn\transpose) \vx + 2\Delta\vn.
        \end{align}
        \item Line reflection: for a line defined by a point $\vp$ and a normal $\vn$, a point $\vx$ is mapped to
        \begin{align}
            \vx' &= R\transpose
            \begin{pmatrix*}[r]
                -1 & 0 & 0\\
                0 & -1 & 0\\
                0 & 0 & \hphantom{-}1
            \end{pmatrix*}
            R (\vx - \vp) + \vp,
        \end{align}
        where the rotation is defined by the implicit equation $\vn = R [0, 0, 1]\transpose$.
        A minimal parametrisation can be constructed by defining the line by a direction vector $\vn$ parallel to the line and a 2D offset $\Delta \in \reals^2$ where the line intersects the plane, in the plane's coordinate system (4 DoF).
        \item Point reflection: for a point $\vp$, a point $\vx$ is mapped to
        \begin{align}
            \vx' &= -(\vx - \vp) + \vp = -\vx +2\vp.
        \end{align}
    \end{enumerate}
    \item \textbf{Rotation symmetry:} whereby a point is mapped to $n$ points (not including the source point) about an arbitrary axis with angle $\theta = 2\pi / (n + 1)$. Thrun \& Wegbreit \cite{thrun2005shape} use the terminology \textit{axial symmetry} to describe the limiting case as $n \to \infty$, such as a cylinder. For an axis defined by a point $\vp$ and a normal $\vn$, a point $\vx$ is mapped to
    \begin{align}
        \vx' &= R\transpose
        \begin{pmatrix*}[r]
            \cos{\theta_k} & \hphantom{-}\sin{\theta_k} & 0\\
            -\sin{\theta_k} & \cos{\theta_k} & 0\\
            0 & 0 & \hphantom{-}1
        \end{pmatrix*}
        R (\vx - \vp) + \vp,
    \end{align}
    where the rotation is defined by the implicit equation $\vn = R [0, 0, 1]\transpose$, and the angle $\theta_k = k\theta$, for $k \in \{1, \dots, n\}$.
    \item \textbf{Spherical symmetry:} whereby a point is mapped to a sphere. For a spherical symmetry with centre $\vp$, a point $\vx$ is mapped to
    \begin{align}
        \vx' &= 
        \begin{pmatrix*}[r]
            1 & 0 & 0\\
            0 & \cos{\alpha} & \hphantom{-}\sin{\alpha}\\
            0 & -\sin{\alpha} & \cos{\alpha}\\
        \end{pmatrix*}
        \begin{pmatrix*}[r]
            \cos{\beta} & 0 & \hphantom{-}\sin{\beta}\\
            0 & \hphantom{-}1 & 0\\
            -\sin{\beta} & 0 & \cos{\beta}
        \end{pmatrix*}
        (\vx - \vp) + \vp,
    \end{align}
    for arbitrary angles $\alpha, \beta$.
    \item \textbf{Translation symmetry:} whereby a point is mapped to another point by applying a fixed translation $t$. For a translation $t$, a point $\vx$ is mapped to $\vx' = \vx + \vt$.
    \item \textbf{Scale symmetry:} whereby a point is mapped to another point by a scaling operation. For scaling factors $s_x$ and $s_y$, a point $\vx$ is mapped to
    \begin{align}
        \vx' &= 
        \begin{pmatrix*}
            s_x & 0 & 0\\
            0 & s_y & 0\\
            0 & 0 & 1
        \end{pmatrix*}
        \vx.
    \end{align}
\end{enumerate}
Composite types can be formed from combinations of these basic types. 
We implement rotational and spherical symmetries in the same way as the bijective symmetries, except that we sample a single point at random from the resulting set, since our approach is point-based. More points could be sampled, at the expense of greater memory and computational requirements.

\paragraph{Multiple symmetries.}
As mentioned in Section 3, our framework has the flexibility to handle multiple symmetries, including different types (\eg, reflections and rotations).
To do so, whenever we apply a symmetry transformation, we choose it at random from the set of pre-defined symmetries.
For example, points sampled in a scene containing a table with two planar reflection symmetries are transformed according to one of those symmetries at random each iteration.
This has the effect of encouraging all of the symmetries in the set without increasing the computational burden.
We can further define relations between the symmetries, such as orthogonality constraints, which limit the degrees of freedom and so facilitate optimisation.

\section{Further Implementation Details}%
\label{sec:implementation}

\subsection{Architecture and Parameters}
\label{sec:architecture}
Following prior art~\cite{yariv2020multiview,wang2021neus}, we implement the SDF network $\phi_\text{SDF}$ as an 8-layer MLP with hidden dimension 256, position-encoded inputs (6 frequencies)~\cite{mildenhall2020nerf}, a skip connection at layer 4, and geometric initialisation for the network weights~\cite{atzmon2020sal}.
The latter provides a spherical prior for the SDF, which we squash into an ellipsoid using the bounding box dimensions.
The material, diffuse, and specular networks are also implemented as MLPs with 4/2/4 hidden layers respectively, with a 4-frequency positional encoding on the normal (diffuse and specular networks) and view (specular network) directions.
The lighting networks have two sets of weights $\vtheta_0$ and $\vtheta_1$ to allow them to model source lighting and symmetry transformed lighting without any symmetry constraint.
NeRF++~\cite{zhang2020nerf} is used as the background model, with default parameters.

We follow the hierarchical sampling strategy of NeuS~\cite{wang2021neus} with 64 coarse, 64 fine, and 32 background samples per ray, with 1024 rays sampled per batch.
The fine samples are obtained using importance sampling, given the density $w$ estimated from the SDF values obtained under the coarse samples (Eq.~3 from the main paper). The peakiness of this distribution is controlled by the learned scalar $\tau$ (initialised to 20), being proportional to the inverse standard deviation of the weight function.
The default hyperparameters are $[\lambdasym, \lambda^\text{d}, \lambda^\text{l}, \lambda^\text{e}] = [0.1, 0.01, 0.001, 0.1]$. 

\subsection{Optimisation}
\label{sec:optimisation}
We optimise the network with Adam~\cite{kingma2014adam} and an initial learning rate of 5e-4, and train for 300k iterations on a single GPU, with an un-optimised implementation taking 31h on a NVIDIA V100. Recent work, such as the multi-resolution hash encoding \cite{mueller2022instant}, is likely to speed up training by orders of magnitude.
We apply cosine annealing to reduce the learning rate gradually over training, reaching $0.05\times\text{LR}$ at the end of training. We also ramp the learning rate of the SDF network, specular lighting network, symmetry parameters, and the inverse variance $\tau$ from 0 at the start of training, for the first 2500 iterations.
On the other hand, the learning rate for the background, the diffuse lighting and the material networks are kept constant at the initial level.
This warm-up period gives the background model a head-start, to avoid modelling the background with the foreground model (\eg, a sky-blue foreground blob above the car can entirely explain the training data in some scenes), and suppresses the specular network, which can otherwise explain away the geometry with arbitrary reflections. The latter only becomes a problem in NeRF-like models when the view density is much lower for some parts of the object.
An alternative approach, which is sometimes able to reconstruct more crisp geometry, is to treat all parameters equally. However, this occasionally fails to correctly model highly-reflective surfaces that are only viewed obliquely, instead deforming the geometry to model the reflections in a process akin to an optical illusion or trompe-l'\oe{}il.

\subsection{Initialisation}
\label{sec:initialisation}
We initialise our coordinate system in the following way. Given a noisy, sparse, and outlier-ridden structure-from-motion (SfM) point cloud, we first filter the point cloud by removing point clusters or singletons further than a threshold (0.2m) from the biggest cluster, and then remove points with fewer than 16 neighbours within the threshold. Next, we fit a 3D bounding box by estimating the up direction from the camera. Then we flatten the point cloud along that dimension, extract the line with greatest support (using a threshold of 0.05) from the 2D point cloud using RANSAC \cite{fischler1981random}, and rotate the point cloud so that the y axis is parallel to the extracted line. The rationale for this approach was that the SfM point clouds often only model one side of the object, and have highly-variable point densities. As a result, extracting the principal direction of the point cloud using a non-robust technique such as eigendecomposition results in poorly aligned bounding boxes.
Composing these transformations, we obtain a new coordinate system and an axis-aligned bounding box.
This is used to initialise the symmetry parameters (\eg, the reflection plane is taken as one of the principal planes), the ground plane (initialised at the height of the bottom of the bounding box), and the SDF ellipsoid prior (the spherical SDF prior is squashed according to the bounding box dimensions).

In our experiments, the 3D point clouds were provided by the benchmark dataset (CO3D~\cite{reizenstein2021common}), which were originally obtained using structure-from-motion (COLMAP~\cite{schoenberger2016sfm}), followed by filtering with PointRend~\cite{kirillov2020pointrend} instance segmentation masks to identify the 3D points on the object.
See the CO3D paper~\cite{reizenstein2021common} for details.

\section{Results on Other CO3D Categories}%
\label{sec:noncar}

\begin{table}[!t]\centering
\caption{
Results for $N$ scenes of the structured test split of the CO3D dataset [24] for \textit{non-car categories}. We report the peak signal-to-noise ratio (PSNR) and LPIPS distance between the estimated and ground-truth masked images, the mean squared error (MSE) between the estimated and ground-truth masked depth maps, and the intersection-over-union (IoU) of the estimated and ground-truth masks.
}\label{tab:co3d_noncar}%
\setlength{\tabcolsep}{0pt}%
\begin{tabularx}{\linewidth}{@{}l c C C C C C C C C@{}}
	\toprule
	&& \multicolumn{2}{c}{PSNR RGB $\uparrow$} & \multicolumn{2}{c}{LPIPS RGB $\downarrow$} & \multicolumn{2}{c}{MSE Depth $\downarrow$} & \multicolumn{2}{c}{IoU Mask $\uparrow$}\\
	Category & $N$ & NeuS & Ours & NeuS & Ours & NeuS & Ours & NeuS & Ours\\
	\midrule
	toyplane & 10 & 
	13.6 & \textbf{15.1} & 0.56 & \textbf{0.51} & 1.18 & \textbf{0.61} & 0.45 & \textbf{0.56}\\
	bench & 8 &
	\textbf{16.0} & 14.8 & \textbf{0.50} & \textbf{0.50} & 0.46 & \textbf{0.41} & 0.43 & \textbf{0.55}\\
	chair & 1 &
	\textbf{18.6} & 16.7 & 0.48 & \textbf{0.47} & 1.01 & \textbf{0.44} & 0.85 & \textbf{0.88}\\
	toytrain & 1 &
	9.60 & \textbf{14.6} & 0.58 & \textbf{0.43} & 0.14 & \textbf{0.09} & \textbf{0.63} & 0.62\\
	toaster & 1 &
	17.8 & \textbf{18.8} & 0.50 & \textbf{0.48} & 3.70 & \textbf{0.05} & 0.82 & \textbf{0.83}\\
	motorcycle & 1 &
	11.0 & \textbf{13.0} & 0.63 & \textbf{0.62} & \textbf{2.30} & 3.88 & 0.70 & \textbf{0.71}\\
	skateboard & 1 &
	11.4 & \textbf{13.0} & \textbf{0.68} & 0.69 & 0.13 & \textbf{0.11} & \textbf{0.20} & \textbf{0.20}\\
	couch & 1 &
	11.5 & \textbf{12.6} & \textbf{0.62} & 0.63 & 1.25 & \textbf{0.32} & \textbf{0.37} & 0.35\\
	parkingmeter & 1 &
	11.4 & \textbf{15.3} & 0.50 & \textbf{0.42} & 0.42 & \textbf{0.36} & 0.82 & \textbf{0.83}\\
	suitcase & 1 &
	13.5 & \textbf{13.7} & \textbf{0.50} & 0.51 & 0.23 & \textbf{0.08} & \textbf{0.80} & \textbf{0.80}\\
	\bottomrule
\end{tabularx}%
\end{table}

\begin{figure}[!t]\centering
\setlength{\tabcolsep}{0pt}
{\scriptsize
\begin{tabularx}{\linewidth}{@{}CCCCC@{}}
\includegraphics[width=0.195\textwidth]{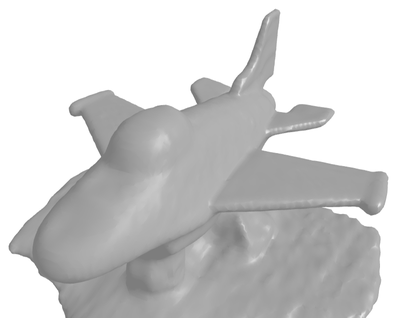} &
\includegraphics[width=0.195\textwidth]{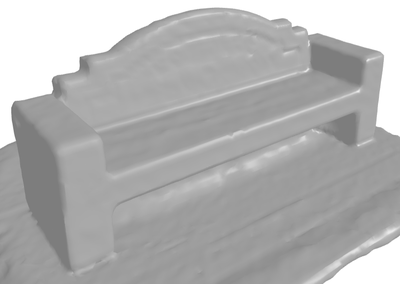} &
\includegraphics[width=0.11\textwidth]{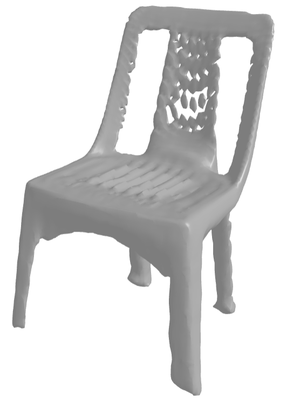} & \includegraphics[width=0.15\textwidth]{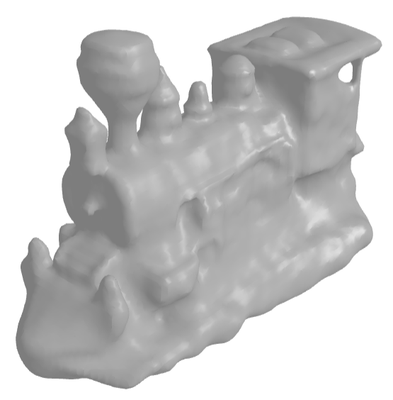} & \includegraphics[width=0.13\textwidth]{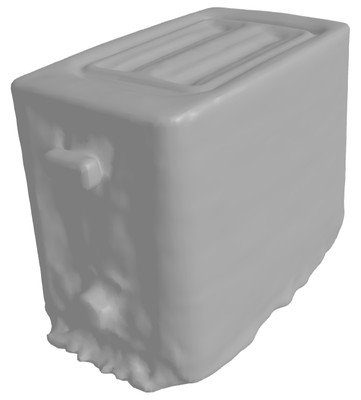}\\
\includegraphics[width=0.195\textwidth]{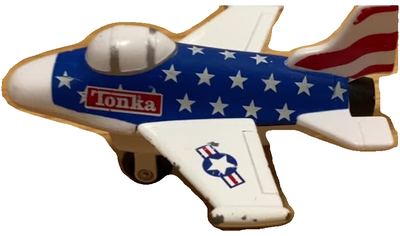} &
\includegraphics[width=0.15\textwidth]{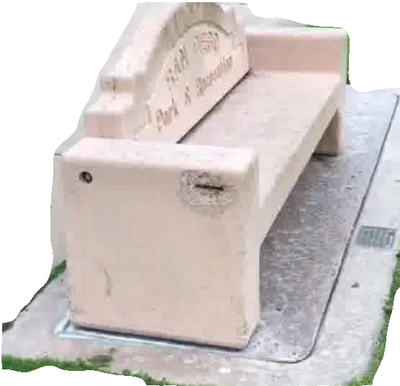} &
\includegraphics[width=0.09\textwidth]{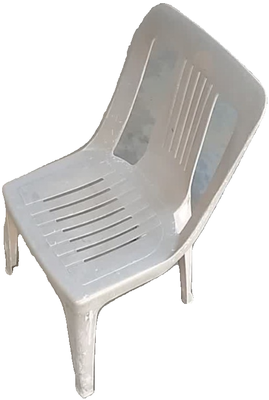} & \includegraphics[width=0.195\textwidth]{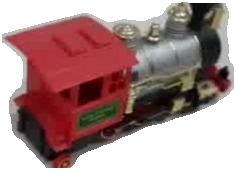} & \includegraphics[width=0.15\textwidth]{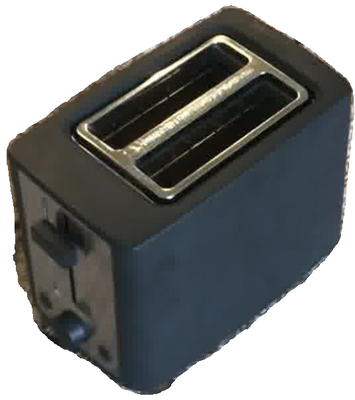}\\
\includegraphics[width=0.195\textwidth]{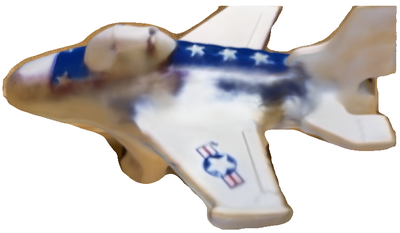} &
\includegraphics[width=0.15\textwidth]{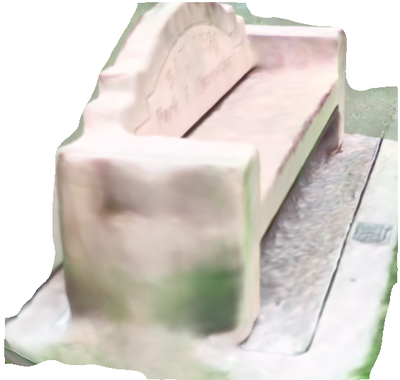} &
\includegraphics[width=0.09\textwidth]{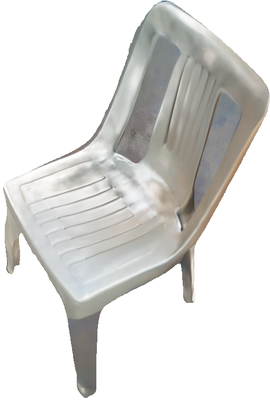} & \includegraphics[width=0.195\textwidth]{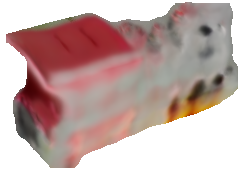} & \includegraphics[width=0.15\textwidth]{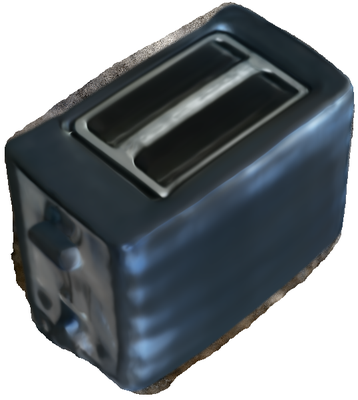}\\
\includegraphics[width=0.195\textwidth]{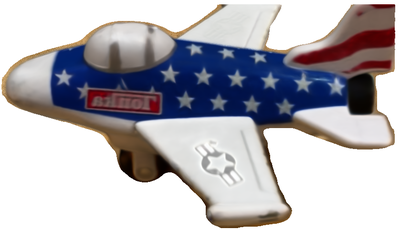} &
\includegraphics[width=0.15\textwidth]{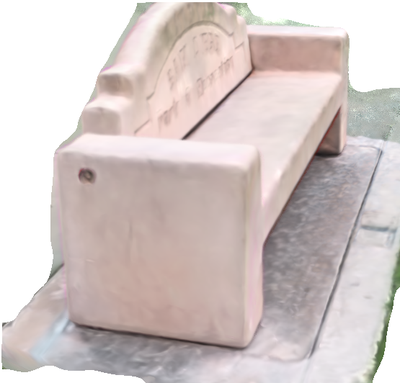} &
\includegraphics[width=0.09\textwidth]{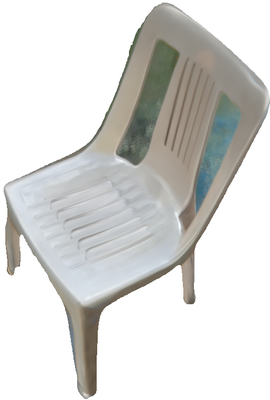} & \includegraphics[width=0.195\textwidth]{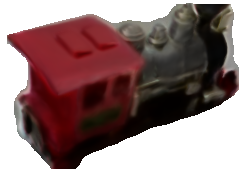} & \includegraphics[width=0.15\textwidth]{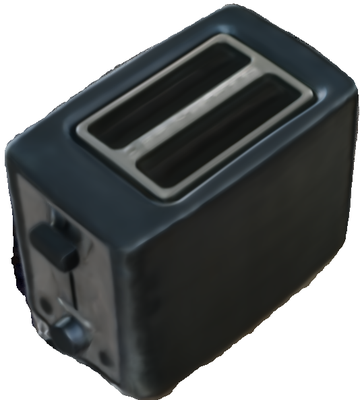}\\
toyplane & bench & chair & toytrain & toaster\\
\includegraphics[width=0.195\textwidth]{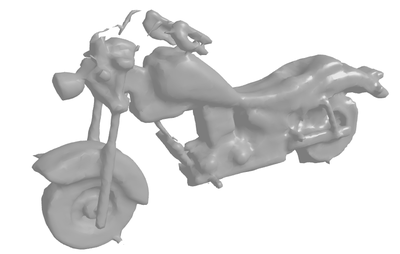} & \includegraphics[width=0.195\textwidth]{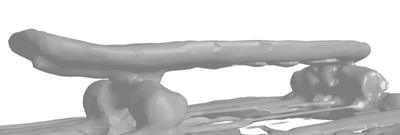} &
\includegraphics[width=0.195\textwidth]{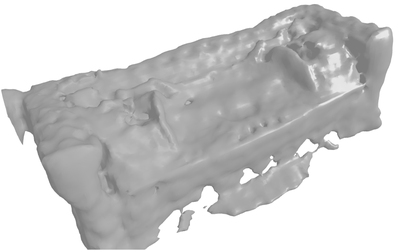} &
\includegraphics[width=0.07\textwidth]{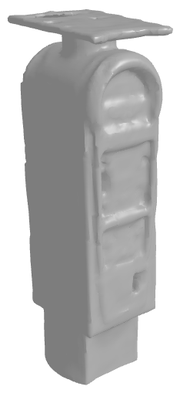} &
\includegraphics[width=0.18\textwidth]{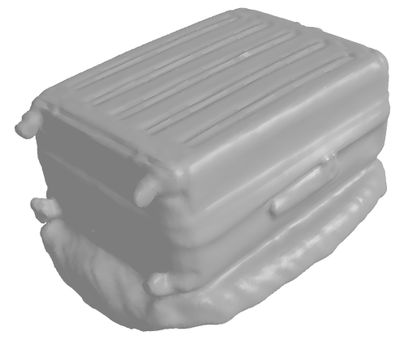}\\
\includegraphics[width=0.18\textwidth]{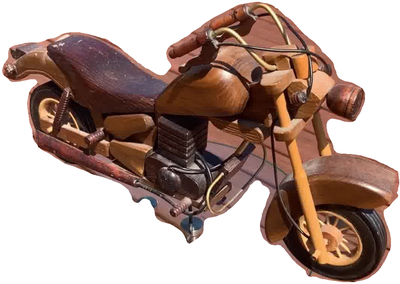} & \includegraphics[width=0.195\textwidth]{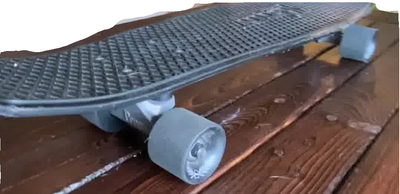} &
\includegraphics[width=0.1\textwidth]{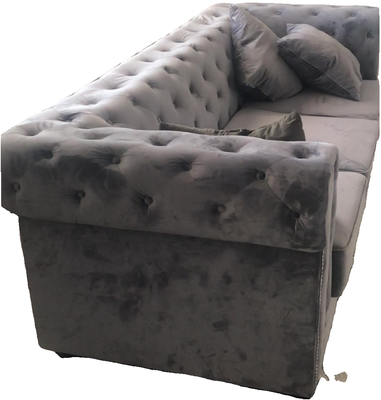} &
\includegraphics[width=0.06\textwidth]{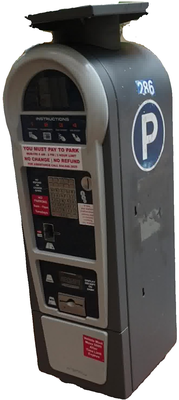} &
\includegraphics[width=0.195\textwidth]{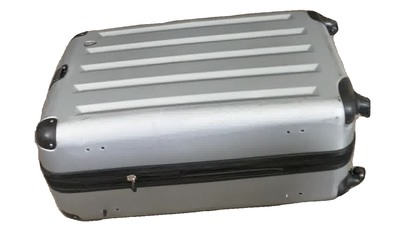} \\
\includegraphics[width=0.18\textwidth]{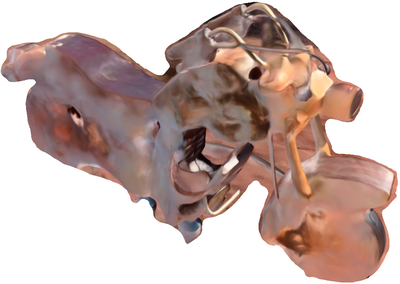} & \includegraphics[width=0.195\textwidth]{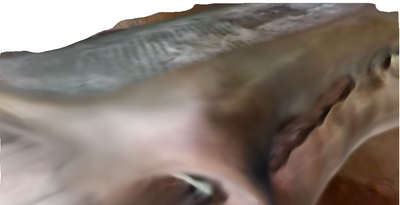} &
\includegraphics[width=0.1\textwidth]{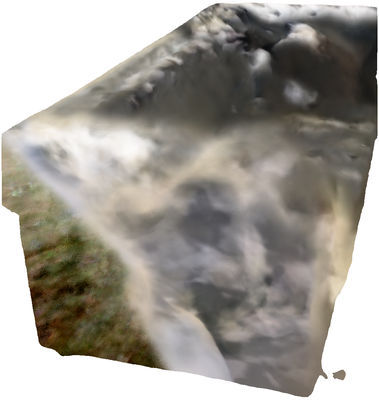} &
\includegraphics[width=0.06\textwidth]{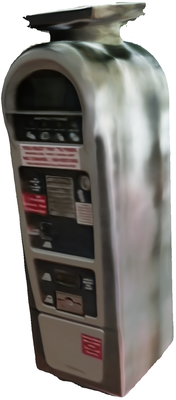} &
\includegraphics[width=0.195\textwidth]{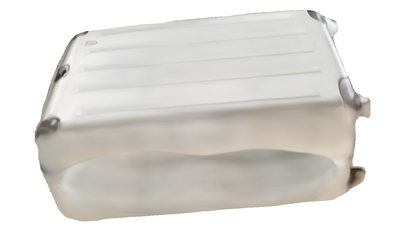} \\
\includegraphics[width=0.18\textwidth]{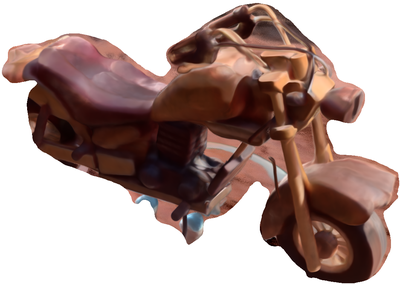} & \includegraphics[width=0.195\textwidth]{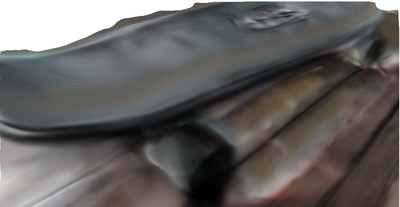} &
\includegraphics[width=0.1\textwidth]{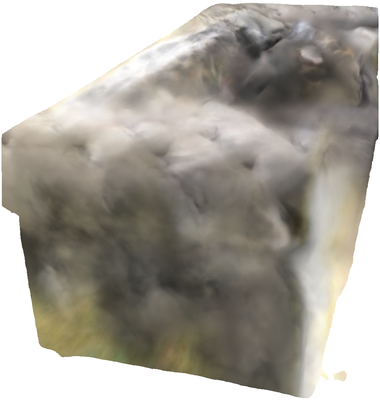} &
\includegraphics[width=0.06\textwidth]{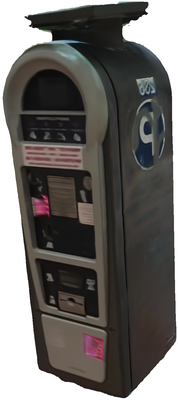} &
\includegraphics[width=0.195\textwidth]{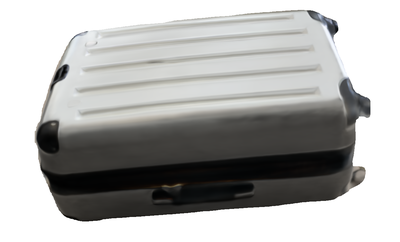} \\
motorcycle & skateboard & couch & parkingmeter & suitcase
\end{tabularx}%
}
\caption{Reconstructions and renders of the unseen sides of 10 non-car categories of the CO3D dataset \cite{reizenstein2021common}. (Rows 1 \& 5) Our reconstruction; (rows 2 \& 2) ground-truth novel view; (rows 3 \& 7) NeuS~\cite{wang2021neus} render; (rows 4 \& 8) our render.}%
\label{fig:noncar}
\vspace{-12pt}
\end{figure}

In this section, we present results on non-car categories from the CO3D dataset~\cite{reizenstein2021common}. We evaluate 10 non-car categories under the challenging setting where we take the minor sector ($130^\circ$) of the structured split as training data and test on the unseen major sector. We test 1 scene per category, except for toyplane and bench, where we evaluated 10 and 8 scenes respectively. The results in \cref{tab:co3d_noncar} show that our method generalises to other symmetric categories. While SNeS always reconstructs the unseen side better than the baseline (NeuS~\cite{wang2021neus}), neither method can predict the unseen lighting. Since our model generates plausible symmetric lighting in the absence of other evidence, this can result in shadow being predicted when the unseen side is under direct sunlight. As a result, our model's PSNR can sometimes be worse than the implausible predictions of the baseline methods on the unseen side. See, for example, the suitcase in \cref{fig:noncar}.

\section{Ablation Study}%
\label{sec:noncar}

\begin{table}[!t]\centering
	\caption{Ablation study on a random subset of our structured test split of the CO3D cars dataset~\cite{reizenstein2021common}.
	We report the peak signal-to-noise ratio (PSNR), mean squared error (MSE), and LPIPS distance between the estimated and ground-truth masked images, the mean absolute error (MAE) between the estimated and ground-truth masked depth maps, and the intersection-over-union (IoU) of the estimated and ground-truth masks.
	The indices $jk$ of the colour losses are 0 for source and 1 for symmetry-transformed geometry/material~($j$) and lighting~($k$).
	}%
	\label{tab:ablation}
\renewcommand*{\arraystretch}{1.1} %
\setlength{\tabcolsep}{0pt} %
\begin{tabularx}{\linewidth}{@{}l C C C C C@{}}
\toprule
Method & PSNR RGB $\uparrow$ & MSE RGB $\downarrow$ & LPIPS RGB $\downarrow$ & MAE Depth $\downarrow$ & IoU Mask $\uparrow$\\
\midrule
SNeS (ours) & \textbf{14.3} & \textbf{0.0372} & 0.564 & 0.0706 & 0.894\\
$+\Ls^\text{lighting}$ & 13.7 & 0.0425 & 0.585 & 0.0685 & 0.914\\
$-\Ls^\text{diffuse}$ & \textbf{14.3} & \textbf{0.0372} & 0.566 & 0.0722 & \textbf{0.917}\\
$-\!$ ground pred. & \textbf{14.3} & \textbf{0.0372} & \textbf{0.558} & \textbf{0.0593} & 0.762\\
$\lambdasym = 1$ & 13.7 & 0.0429 & 0.571 & 0.0627 & 0.873\\
$-\{\Ls_{01}^\text{col}, \Ls_{11}^\text{col}\}$ & 13.9 & 0.0408 & 0.587 & 0.0717 & 0.864\\
$-\{\Ls_{01}^\text{col}, \Ls_{10}^\text{col}\}$ & 13.6 & 0.0437 & 0.561 & 0.0732 & 0.913\\
$-\{\Ls_{01}^\text{col}, \Ls_{10}^\text{col}, \Ls_{11}^\text{col}\}$ & 13.7 & 0.0422 & 0.576 & 0.0782 & 0.906\\
\bottomrule
\end{tabularx}%
\end{table}

In this section, we present an expanded ablation study.
To investigate the effect of the different components of our model, we ablate its performance on a subset of 4 randomly selected scenes from our structured test split of the CO3D cars dataset, as shown in \cref{tab:ablation} and \cref{fig:ablation}.
We ablate with respect to the model without the lighting loss (ours), since this loss is designed to produce qualitatively convincing renders in the absence of image evidence, but is unlikely to be quantitatively (visually) accurate in those areas.
We indeed see that the symmetric lighting loss has a detrimental effect on the image-based results, predominantly in situations where direct sunlight is applied to the shadowed side of the car and vice versa.
However, the resulting renders are qualitatively preferable, and exhibit better geometry.
We verify that removing the diffuse colour loss harms the geometry, since it helps decouple the symmetric and asymmetric properties facilitating symmetry learning.
The ablation of the symmetry losses $\Ls_{jk}^\text{col}$ indicates that each term, including the mixed losses, are important for visual and geometric quality.
In particular, removing all symmetry terms has a significantly detrimental effect on performance, as does over-symmetrising the scene ($\lambdasym = 1$).
An accompanying visual ablation study is given in \cref{sec:ablation_qualitative}.

\section{Additional Qualitative Results}%
\label{sec:qualitative}

\begin{figure}[!t]\centering
    \setlength{\tabcolsep}{0pt}
    \begin{tabular}{cccc}
    \includegraphics[height=0.175\linewidth]{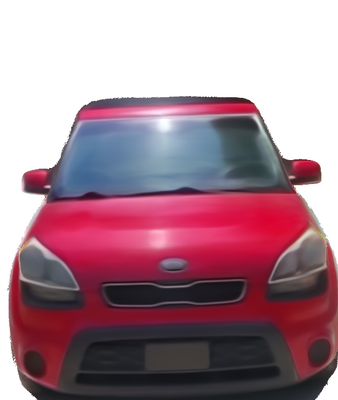} &
    \includegraphics[height=0.175\linewidth]{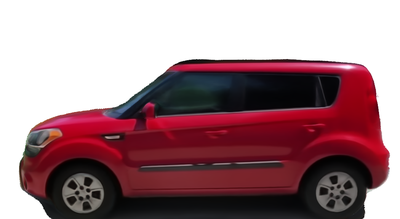} &
    \includegraphics[height=0.175\linewidth]{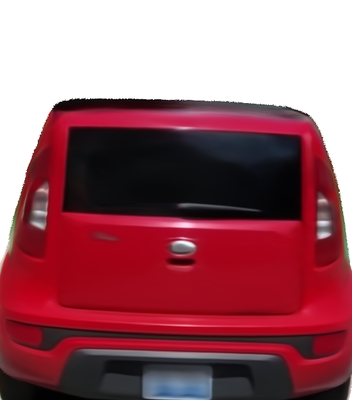} &
    \includegraphics[height=0.175\linewidth]{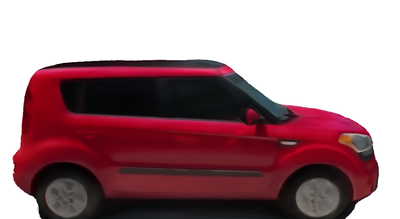}\\[-3.5pt]
    \includegraphics[height=0.175\linewidth]{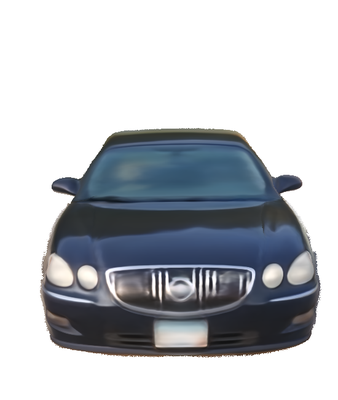} &
    \includegraphics[height=0.175\linewidth]{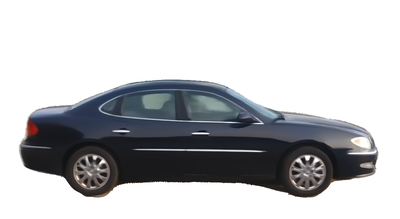} &
    \includegraphics[height=0.175\linewidth]{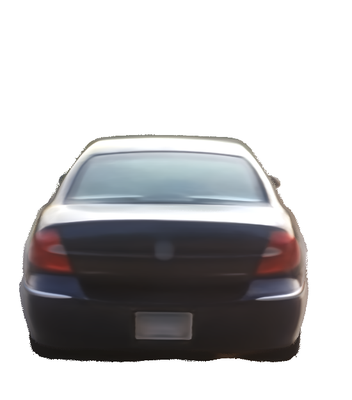} &
    \includegraphics[height=0.175\linewidth]{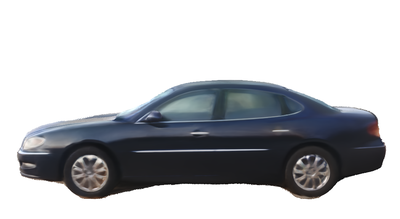}\\[-3.5pt]
    \includegraphics[height=0.175\linewidth]{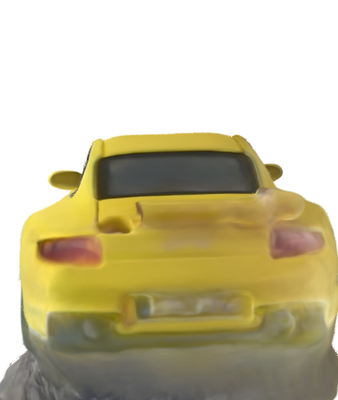} &
    \includegraphics[height=0.175\linewidth]{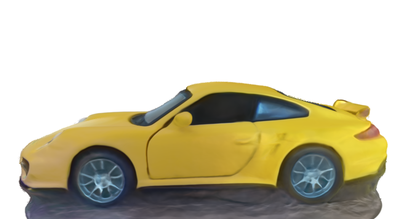} &
    \includegraphics[height=0.175\linewidth]{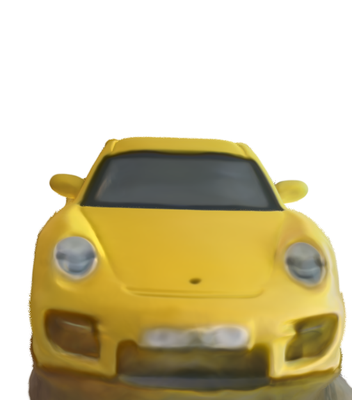} &
    \includegraphics[height=0.175\linewidth]{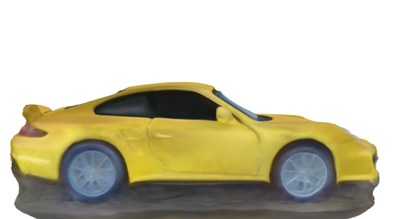}\\[-3.5pt]
    \includegraphics[height=0.175\linewidth]{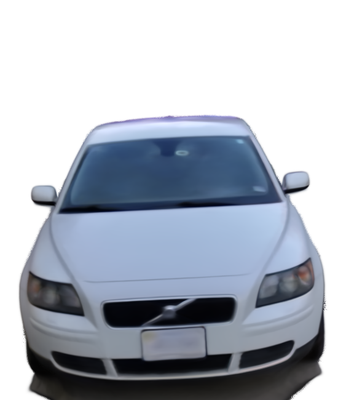} &
    \includegraphics[height=0.175\linewidth]{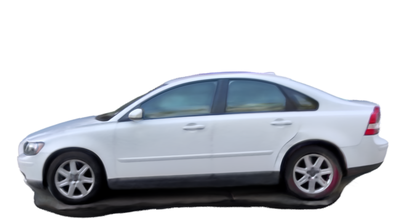} &
    \includegraphics[height=0.175\linewidth]{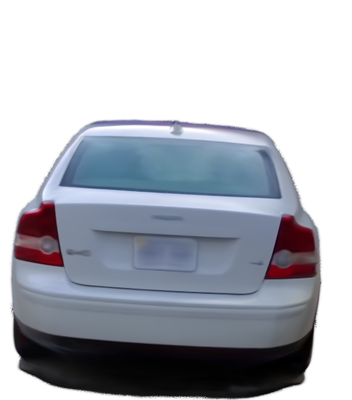} &
    \includegraphics[height=0.175\linewidth]{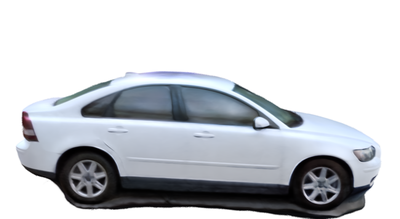}\\[-3.5pt]
    \includegraphics[height=0.175\linewidth]{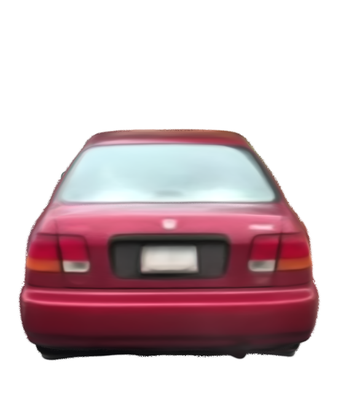} &
    \includegraphics[height=0.175\linewidth]{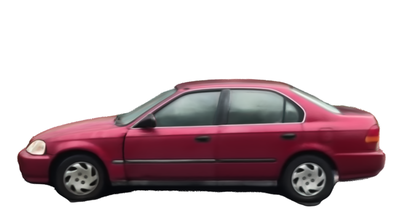} &
    \includegraphics[height=0.175\linewidth]{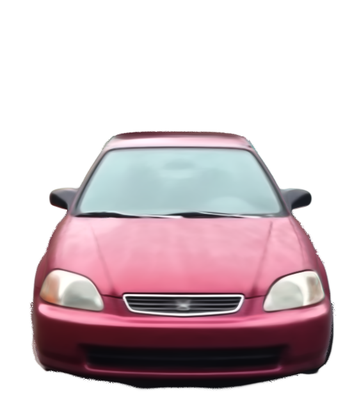} &
    \includegraphics[height=0.175\linewidth]{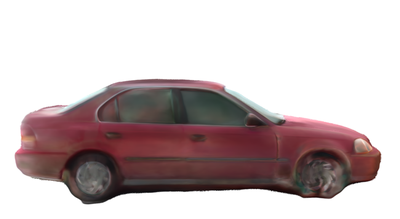}\\[-3.5pt]
    \includegraphics[height=0.175\linewidth]{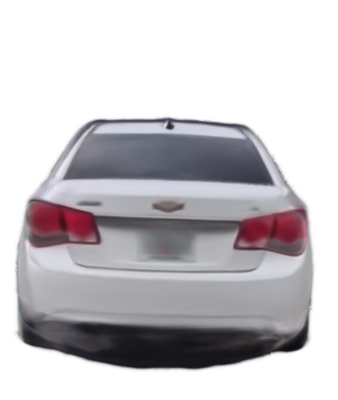} &
    \includegraphics[height=0.175\linewidth]{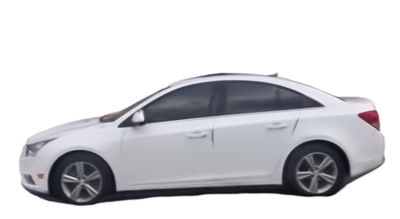} &
    \includegraphics[height=0.175\linewidth]{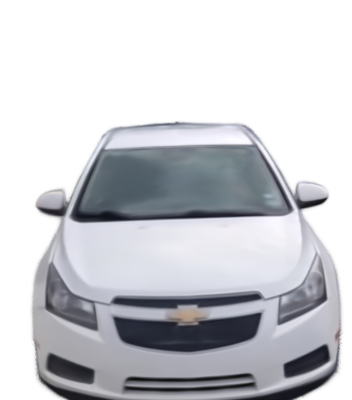} &
    \includegraphics[height=0.175\linewidth]{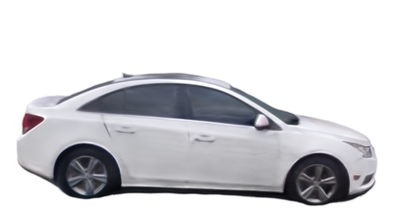}\\[-3.5pt]
    \includegraphics[height=0.175\linewidth]{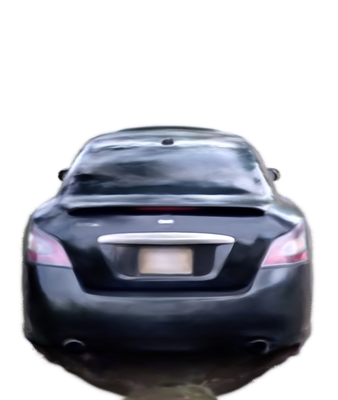} &
    \includegraphics[height=0.175\linewidth]{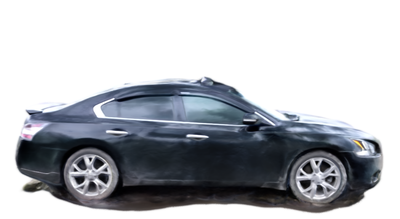} &
    \includegraphics[height=0.175\linewidth]{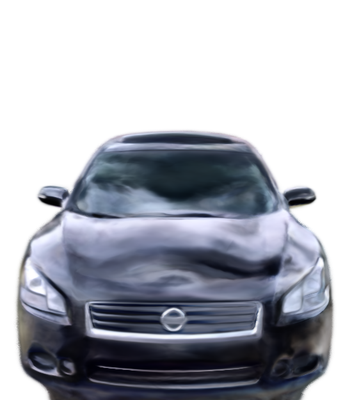} &
    \includegraphics[height=0.175\linewidth]{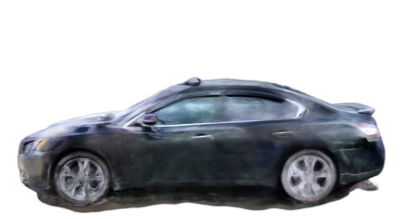}\\[-3.5pt]
    \includegraphics[height=0.175\linewidth]{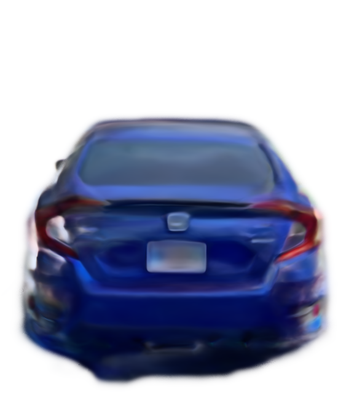} &
    \includegraphics[height=0.175\linewidth]{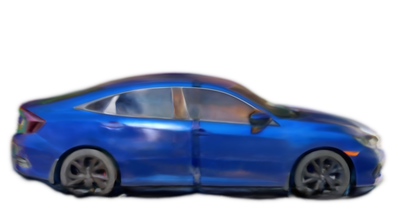} &
    \includegraphics[height=0.175\linewidth]{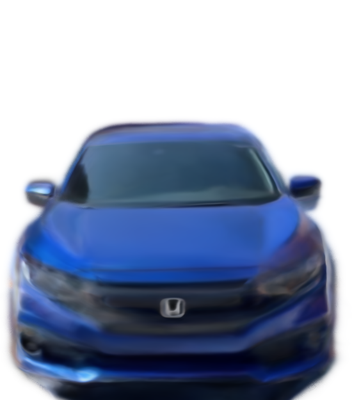} &
    \includegraphics[height=0.175\linewidth]{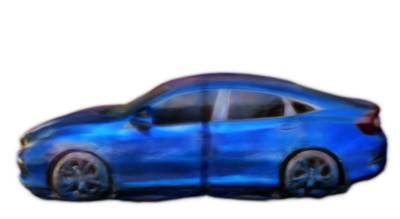}\\[-3.5pt]
    \end{tabular}
    \caption{Additional qualitative results on the car category of the CO3D dataset \cite{reizenstein2021common}. The last column shows the side of the car unseen during training.}%
    \label{fig:extra_qualitative}
\end{figure}

\begin{figure}[!t]\centering
    \setlength{\tabcolsep}{0pt}
    \begin{tabular}{cc}
    \includegraphics[height=0.30\linewidth]{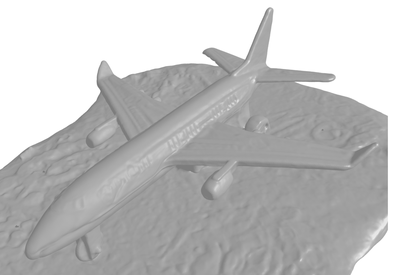} &
    \includegraphics[height=0.30\linewidth]{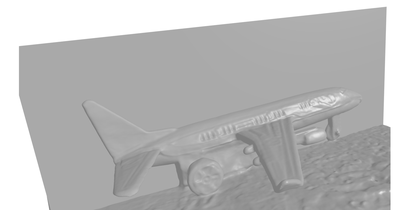} \\
    
    \includegraphics[height=0.35\linewidth]{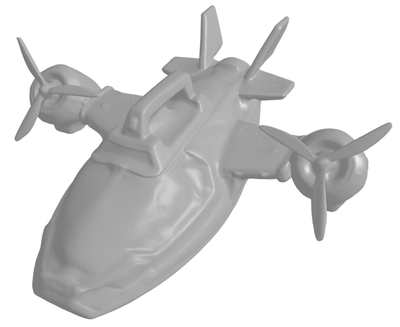} &
    \includegraphics[height=0.35\linewidth]{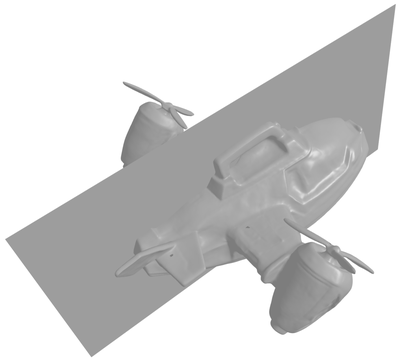}\\[-3.5pt]
    
    \includegraphics[height=0.35\linewidth]{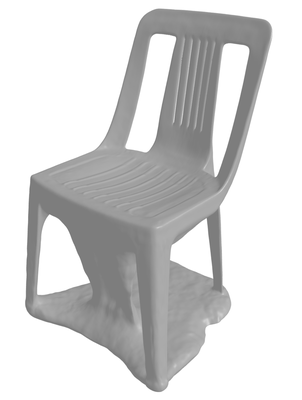} &
    \includegraphics[height=0.35\linewidth]{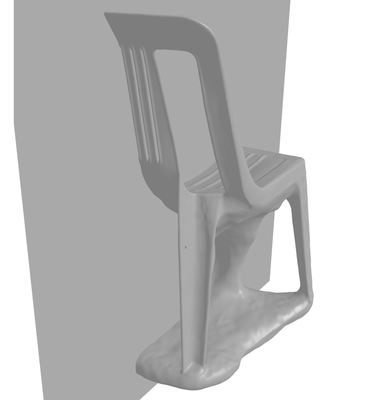} \\

    \includegraphics[height=0.30\linewidth]{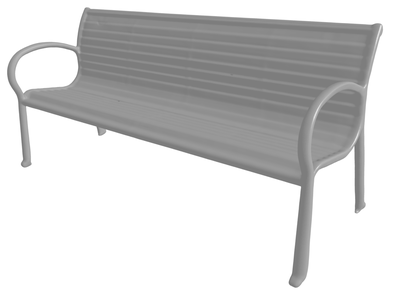} &
    \includegraphics[height=0.30\linewidth]{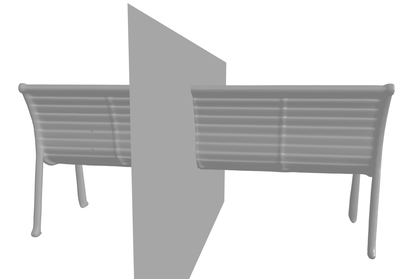} \\
    
    \end{tabular}
    \caption{Reconstructions on additional categories of the CO3D dataset exhibiting reflection symmetry: toyplane (rows 1-2), chair (row 3) and bench (row 4). The second column shows the plane of symmetry.}%
    \label{fig:additional_categories}
\end{figure}

In this section, we present additional high-resolution qualitative results on the CO3D dataset \cite{reizenstein2021common}. We also present a visual decomposition of the material and lighting properties that our model learns, as well as a visual ablation of the components of our model.
We also include high-resolution videos of our novel view synthesis results alongside this document, for the \textit{structured} (\ie partial-view) test split. They clearly show a consistent reconstruction with smoothly varying view-dependent effects and high-frequency specular highlights appearing where appropriate, despite one side of the car being only obliquely seen in the training data.

\subsection{Additional Results on the CO3D Car Dataset}%
\label{sec:car}

In this section, we present additional high-resolution qualitative results on the challenging car category of the CO3D dataset \cite{reizenstein2021common}, shown in \cref{fig:extra_qualitative}. We are able to recover high-fidelity details across the category, despite using real-world data of highly-reflective and texture-poor objects. The bottom two rows show failure cases where the model was not able to adequately deal with the extreme reflections on the car body.

\subsection{Additional Results on Other CO3D Categories}%
\label{sec:categories}

In this section, we demonstrate our model's effectiveness on other near-symmetric categories of the CO3D dataset \cite{reizenstein2021common}, see \cref{fig:additional_categories}. Our method produces high quality reconstructions while accurately estimating the symmetry plane. However, the symmetry parameters do require a reasonable initialisation, as previously discussed, for the local optimisation to converge. Our heuristic (\cref{sec:initialisation}) for estimating the initial alignment exploits the geometric property of cars that their longest dimension is parallel to the symmetry plane. This heuristic could be extended to exhaustively search for the alignment closest to the assumed symmetry plane. Instead, we manually select, for each category, the coordinate axis that is normal to the symmetry plane. For example, the heuristic-extracted axis of the airplane category is the wing axis, which is perpendicular to the symmetry plane. In future work, we intend to predict the symmetry plane parameters directly from the posed images or point cloud, and use this to initialise our model.

\subsection{Appearance Factorisation}%
\label{sec:factorisation}

\begin{figure}[!t]\centering
    \setlength{\tabcolsep}{0pt}
    \begin{tabular}{lcccc}
    \rotatebox[origin=l]{90}{\scriptsize Albedo} &
    \includegraphics[height=0.175\linewidth]{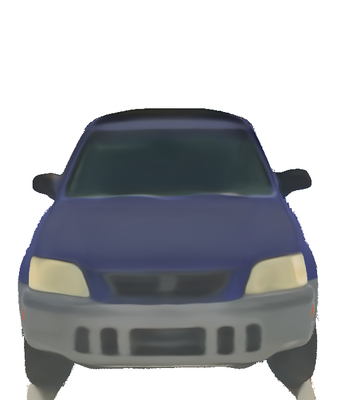} &
    \includegraphics[height=0.175\linewidth]{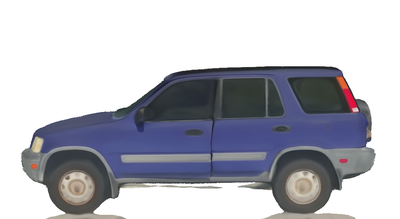} &
    \includegraphics[height=0.175\linewidth]{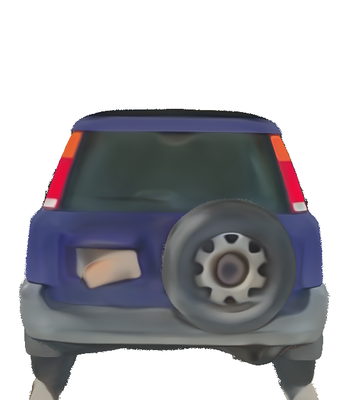} &
    \includegraphics[height=0.175\linewidth]{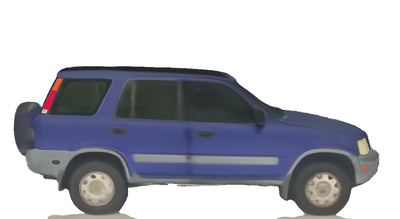}\\[-3.5pt]
    \rotatebox[origin=l]{90}{\scriptsize Alb.$\times$Diff.} &
    \includegraphics[height=0.175\linewidth]{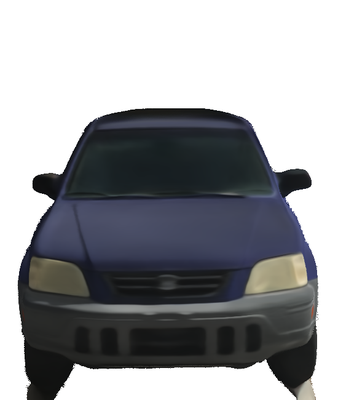} &
    \includegraphics[height=0.175\linewidth]{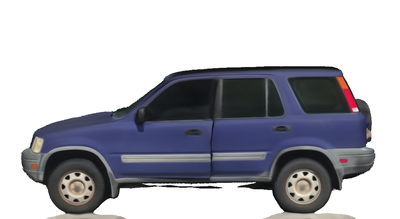} &
    \includegraphics[height=0.175\linewidth]{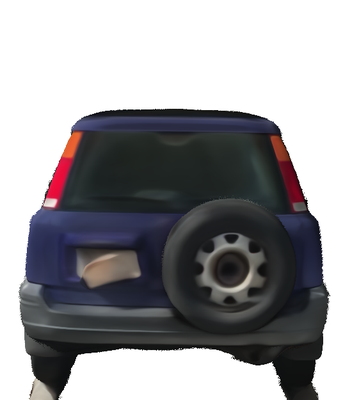} &
    \includegraphics[height=0.175\linewidth]{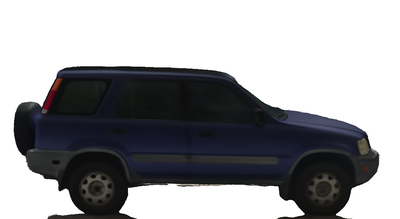}\\[-3.5pt]
    \rotatebox[origin=l]{90}{\scriptsize Specular} &
    \includegraphics[height=0.175\linewidth]{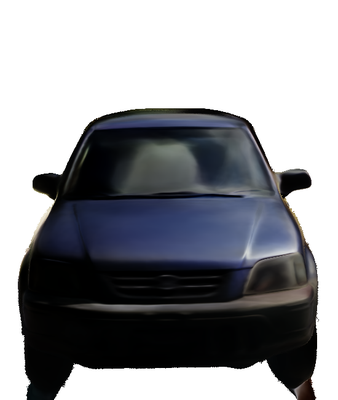} &
    \includegraphics[height=0.175\linewidth]{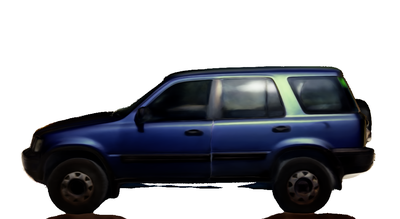} &
    \includegraphics[height=0.175\linewidth]{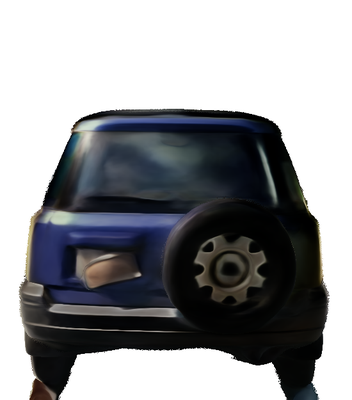} &
    \includegraphics[height=0.175\linewidth]{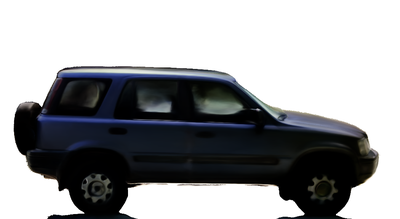}\\[-3.5pt]
    \rotatebox[origin=l]{90}{\scriptsize Spec.$\times$Refl.} &
    \includegraphics[height=0.175\linewidth]{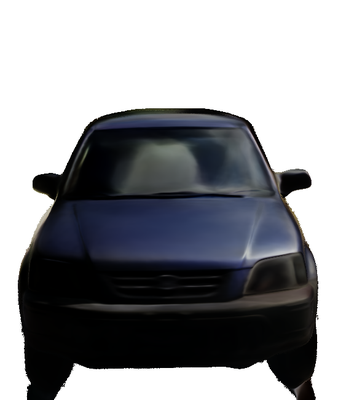} &
    \includegraphics[height=0.175\linewidth]{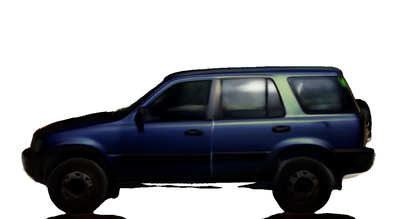} &
    \includegraphics[height=0.175\linewidth]{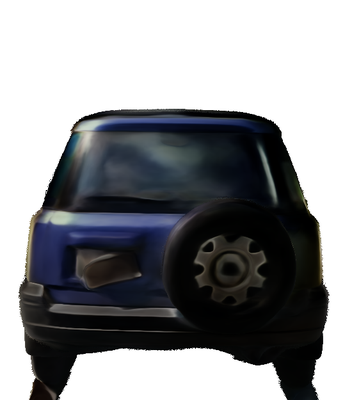} &
    \includegraphics[height=0.175\linewidth]{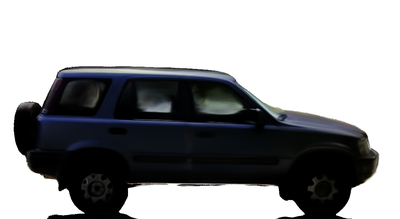}\\[-3.5pt]
    \rotatebox[origin=l]{90}{\scriptsize Full} &
    \includegraphics[height=0.175\linewidth]{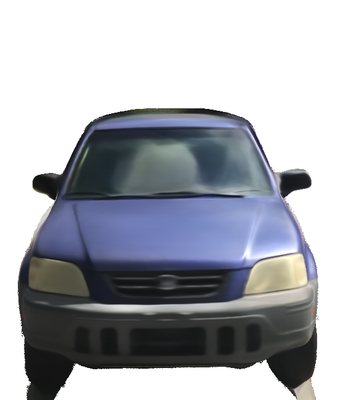} &
    \includegraphics[height=0.175\linewidth]{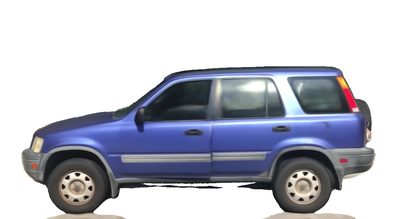} &
    \includegraphics[height=0.175\linewidth]{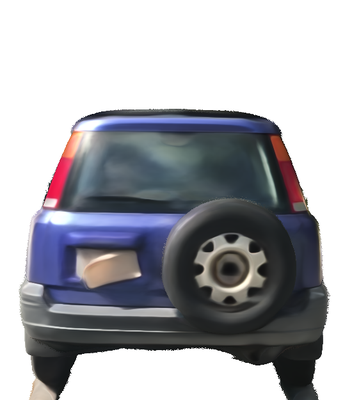} &
    \includegraphics[height=0.175\linewidth]{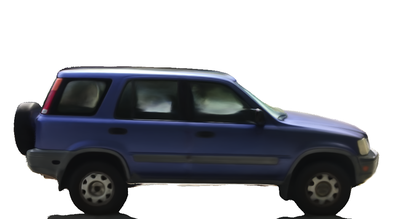}\\[-3.5pt]
    \rotatebox[origin=l]{90}{\scriptsize Ground-truth} &
    \includegraphics[height=0.145\linewidth]{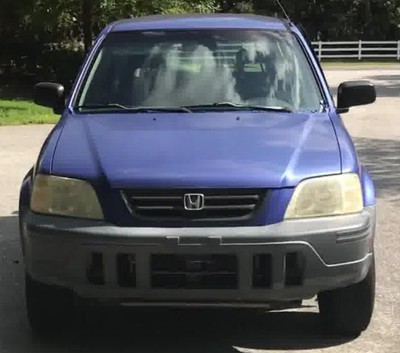} &
    \includegraphics[height=0.145\linewidth]{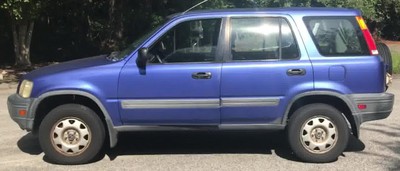} &
    \includegraphics[height=0.145\linewidth]{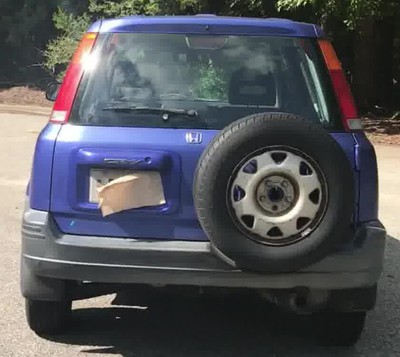} &
    \includegraphics[height=0.145\linewidth]{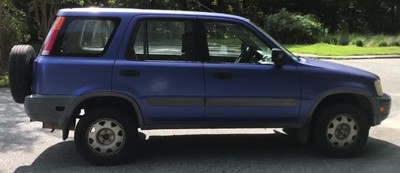}\\[-3.5pt]
    \end{tabular}
    \caption{Visualisation of colour decomposition. Note that, despite significant differences in illumination on the left and right sides of the car, our model recovers a plausible and symmetric albedo.}%
    \label{fig:colour_decomposition}
\end{figure}

In this section, we render images from the factorised components of our colour model, to demonstrate the effectiveness of the decomposition. We show the albedo colour, diffuse shading, reflectivity, and specular colour separately, as well as the diffuse colour (albedo colour $\times$ diffuse shading) and the specular lighting (reflectivity $\times$ specular colour), in \cref{fig:colour_decomposition}. We observe that the colour model is able to disentangle the different components very effectively. In particular, despite significant differences in illumination on the left and right sides of the car, our model is able to recover a plausible and symmetric albedo.

\subsection{Qualitative Ablation Study}%
\label{sec:ablation_qualitative}

\begin{figure}[!t]\centering
    \setlength{\tabcolsep}{0pt}
    \begin{tabular}{lcccc}
    \rotatebox[origin=l]{90}{\tiny Ground-truth} &
    \includegraphics[width=0.25\linewidth]{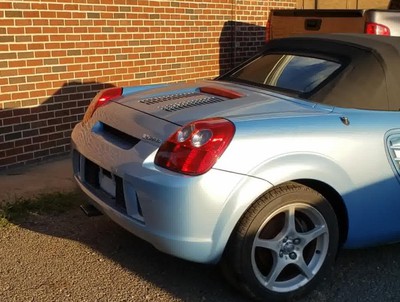} &
    \includegraphics[width=0.25\linewidth]{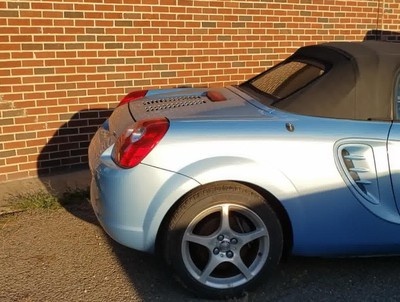} &
    \includegraphics[width=0.25\linewidth]{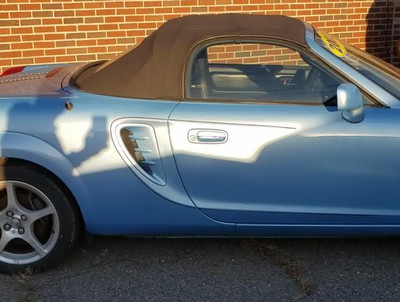} &
    \includegraphics[width=0.25\linewidth]{ablation/gt/rgb/frame000080}\\[-3.5pt]
    \rotatebox[origin=l]{90}{\tiny Ours} &
    \includegraphics[width=0.25\linewidth]{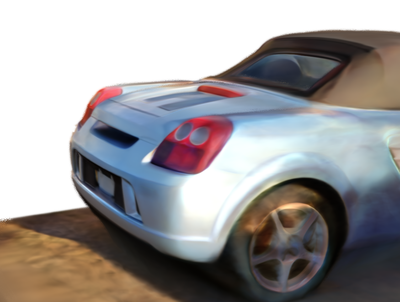} &
    \includegraphics[width=0.25\linewidth]{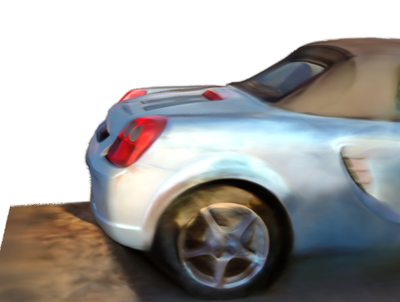} &
    \includegraphics[width=0.25\linewidth]{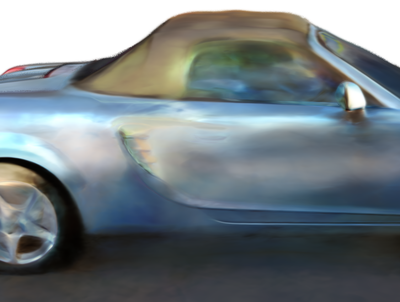} &
    \includegraphics[width=0.25\linewidth]{ablation/wo_symlight/rgb/frame000080}\\[-3.5pt]
    \rotatebox[origin=l]{90}{\tiny $+\Ls^\text{lighting}$} &
    \includegraphics[width=0.25\linewidth]{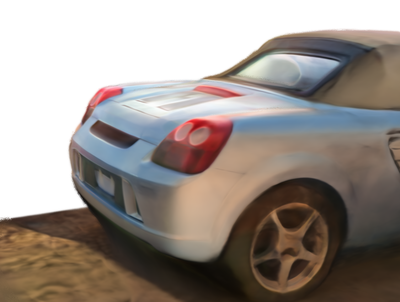} &
    \includegraphics[width=0.25\linewidth]{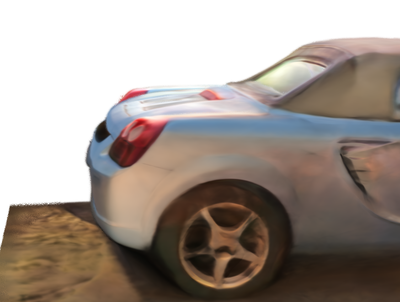} &
    \includegraphics[width=0.25\linewidth]{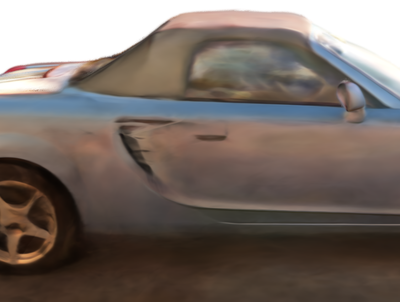} &
    \includegraphics[width=0.25\linewidth]{ablation/ours/rgb/frame000080}\\[-3.5pt]
    \rotatebox[origin=l]{90}{\tiny $-\Ls^\text{diffuse}$} &
    \includegraphics[width=0.25\linewidth]{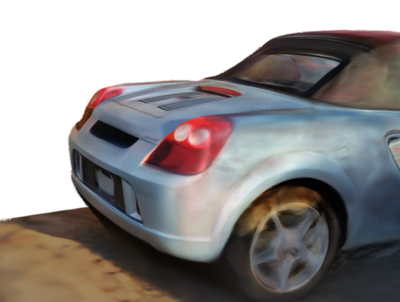} &
    \includegraphics[width=0.25\linewidth]{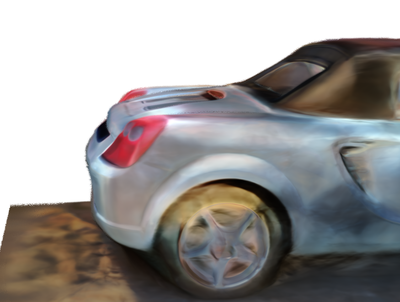} &
    \includegraphics[width=0.25\linewidth]{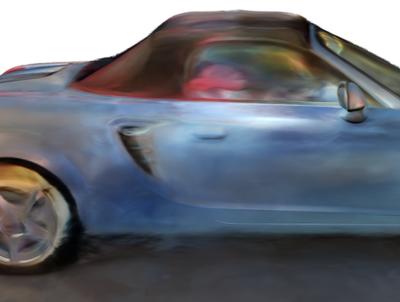} &
    \includegraphics[width=0.25\linewidth]{ablation/wo_diffuse/rgb/frame000080}\\[-3.5pt]
    \rotatebox[origin=l]{90}{\tiny $\lambdasym = 1$} &
    \includegraphics[width=0.25\linewidth]{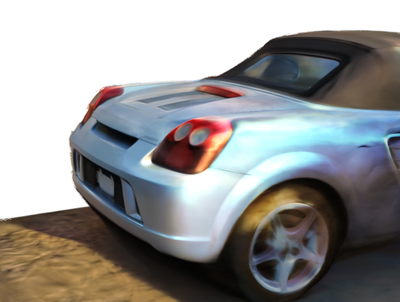} &
    \includegraphics[width=0.25\linewidth]{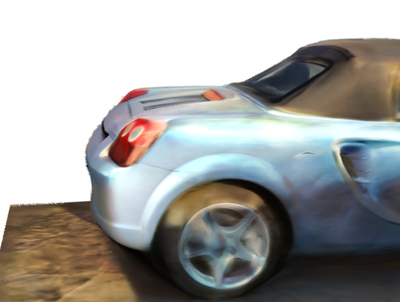} &
    \includegraphics[width=0.25\linewidth]{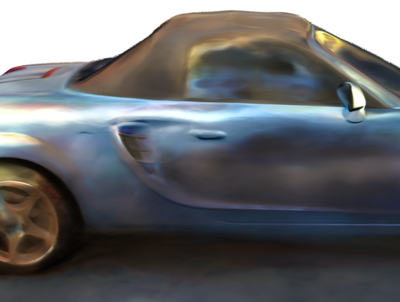} &
    \includegraphics[width=0.25\linewidth]{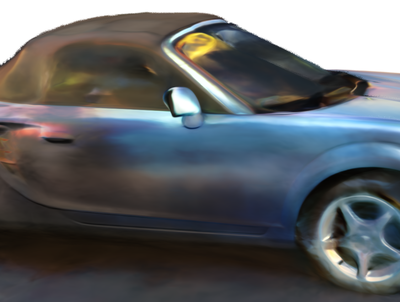}\\[-3.5pt]
    \rotatebox[origin=l]{90}{\tiny $-\{\Ls_{01}^\text{col}\!, \Ls_{11}^\text{col}\}$} &
    \includegraphics[width=0.25\linewidth]{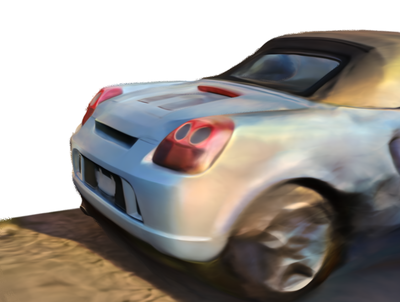} &
    \includegraphics[width=0.25\linewidth]{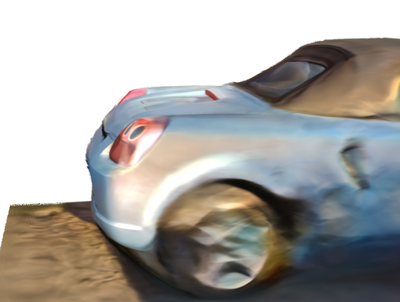} &
    \includegraphics[width=0.25\linewidth]{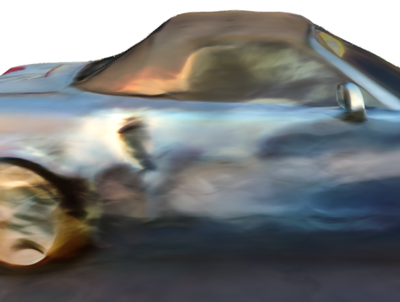} &
    \includegraphics[width=0.25\linewidth]{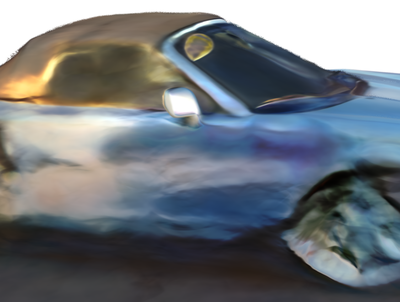}\\[-3.5pt]
    \rotatebox[origin=l]{90}{\tiny $-\{\Ls_{01}^\text{col}\!, \Ls_{10}^\text{col}\}$} &
    \includegraphics[width=0.25\linewidth]{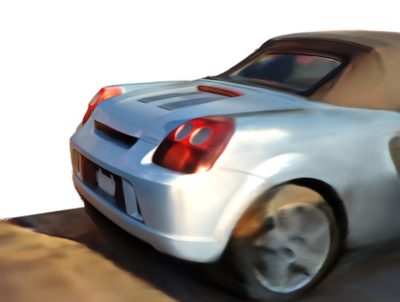} &
    \includegraphics[width=0.25\linewidth]{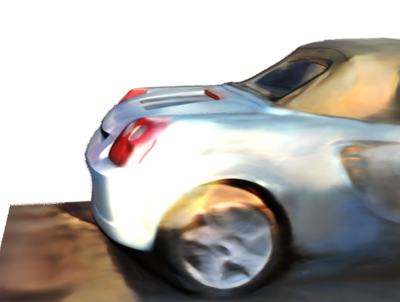} &
    \includegraphics[width=0.25\linewidth]{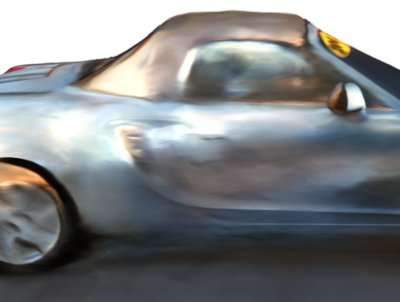} &
    \includegraphics[width=0.25\linewidth]{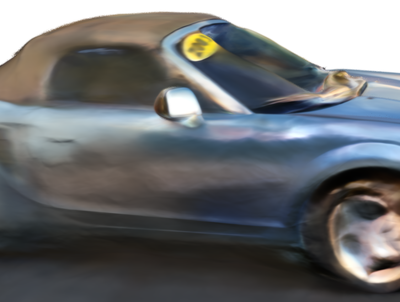}\\[-3.5pt]
    \rotatebox[origin=l]{90}{\tiny $-\{\Ls_{01}^\text{col}\!, \!\Ls_{10}^\text{col}\!, \!\Ls_{11}^\text{col}\}$} &
    \includegraphics[width=0.25\linewidth]{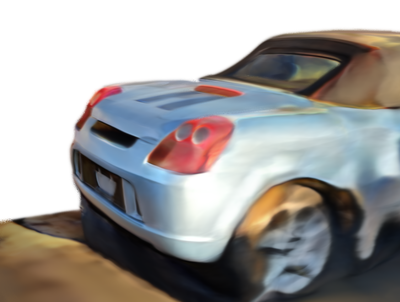} &
    \includegraphics[width=0.25\linewidth]{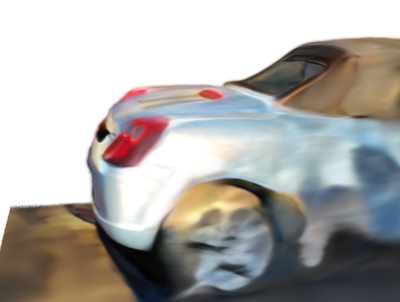} &
    \includegraphics[width=0.25\linewidth]{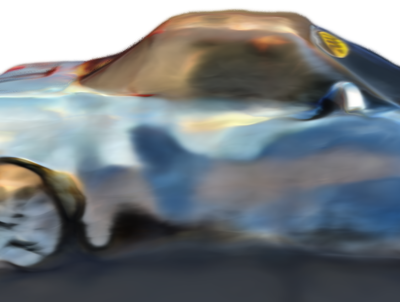} &
    \includegraphics[width=0.25\linewidth]{ablation/wo_011011/rgb/frame000080}
    \end{tabular}
    \caption{Qualitative ablation study. Novel view renderings of the unseen side of the car.
    }%
    \label{fig:ablation}
\end{figure}

In this section, we visually ablate the model, showing the unseen side of one of the ablation sequences. This is a particularly difficult sequence, with strongly asymmetric lighting and moving shadows cast by the cameraperson. We show the results in \cref{fig:ablation}.
We observe that adding the symmetric lighting loss (third row) improves the visual quality of the renders, at the cost of over-symmetrising the lighting.
In comparison, the base model without symmetric lighting (second row) estimates the diffuse lighting reasonably well, but is unable to recover the specular lighting, since these viewing directions are unobserved. A symmetric prior on the specular lighting would be helpful in this case, or a null prior.

Removing the diffuse lighting loss makes it more difficult for the model to disentangle the material, diffuse lighting and specular lighting components. This is particularly visible at the wheels, where the lighting is (incorrectly) modelled as specular reflections.
A higher symmetry factor $\lambdasym = 1$ is quite helpful for this particular scene, since the geometry and (most of) the appearance is symmetric. With a higher weight on the symmetry terms, more detail is transferred from the visible side.
It is also notable that, in the absence of contrary information, the default parameters ($\lambdasym = 0.1$) symmetrise the wheels, which are rotationally offset from the ground truth.

The ablation of the symmetric colour losses are shown in the final three rows. The indices $jk$ of the colour losses are 0 for source and 1 for symmetry-transformed geometry/material~($j$) and lighting~($k$). We see that removing any of these components significantly damages the view synthesis quality, especially removing all symmetry terms (last row).

\begin{figure}[!t]\centering
	\begin{subfigure}[]{0.5\textwidth}\centering
	   \includegraphics[width=\textwidth]{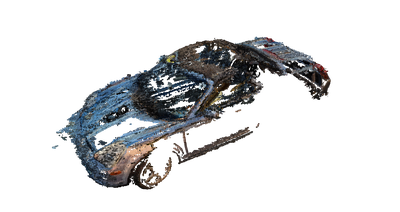}
	   \caption{Ground-truth point cloud (left)}
		\label{fig:pointcloud_1}
	\end{subfigure}\hfill
	\begin{subfigure}[]{0.5\textwidth}\centering
		\includegraphics[width=\textwidth]{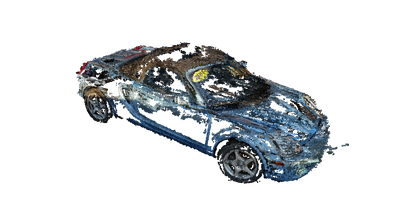}
		\caption{Ground-truth point cloud (right)}
		\label{fig:pointcloud_2}
	\end{subfigure}\vfill
	\begin{subfigure}[]{0.45\textwidth}\centering
	   \includegraphics[width=\textwidth]{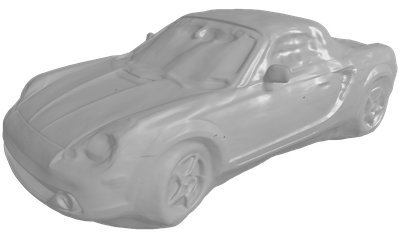}
		\label{fig:pointcloud_3}
	\end{subfigure}\hfill
	\begin{subfigure}[]{0.45\textwidth}\centering
		\includegraphics[width=\textwidth]{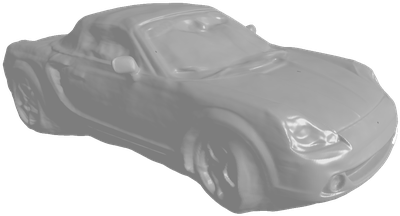}
		\label{fig:pointcloud_4}
	\end{subfigure}\vfill
	\begin{subfigure}[]{0.45\textwidth}\centering
	   \includegraphics[width=\textwidth]{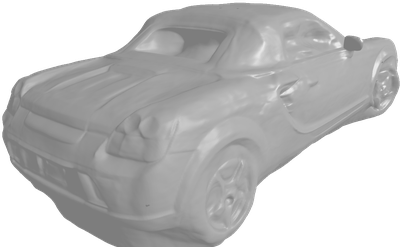}
	   \caption{With symmetry (left)}
		\label{fig:pointcloud_5}
	\end{subfigure}\hfill
	\begin{subfigure}[]{0.45\textwidth}\centering
		\includegraphics[width=\textwidth]{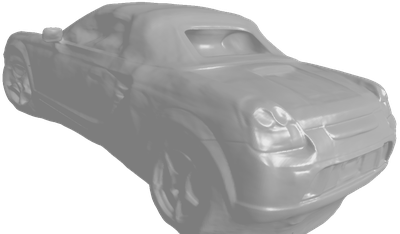}
		\caption{With symmetry (right)}
		\label{fig:pointcloud_6}
	\end{subfigure}
	\begin{subfigure}[]{0.45\textwidth}\centering
	   \includegraphics[width=\textwidth]{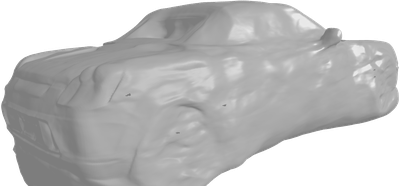}
	   \caption{Without symmetry (left)}
	   \label{fig:pointcloud_7}
	\end{subfigure}\hfill
	\begin{subfigure}[]{0.45\textwidth}\centering
		\includegraphics[width=\textwidth]{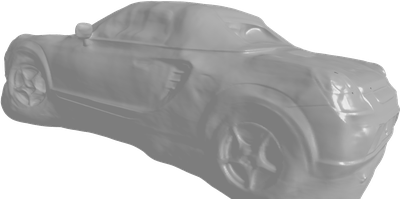}
		\caption{Without symmetry (right)}
		\label{fig:pointcloud_8}
	\end{subfigure}
    \caption{Ground-truth point cloud (top), our reconstruction (middle), and reconstruction without symmetry (bottom) of this ablated sequence from the CO3D dataset~\cite{reizenstein2021common}. The ground-truth is very sparse and noisy, with incorrect geometry in numerous places. In contrast, our reconstruction is dense and detailed, able to reconstruct texture-poor regions despite their high reflectivity. When the symmetry terms are not used, the model fails to reconstruct any detail on the unseen side.}%
    \label{fig:ablation_pointcloud}
\end{figure}

\begin{figure}[!t]\centering
    \setlength{\tabcolsep}{0pt}
    \begin{tabular}{cccc}
    \includegraphics[width=0.25\linewidth]{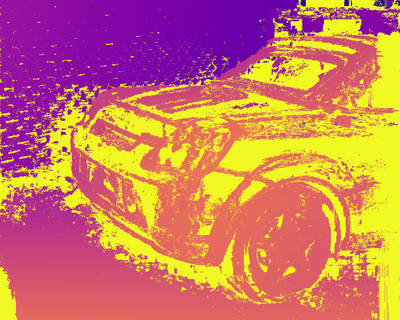} &
    \includegraphics[width=0.25\linewidth]{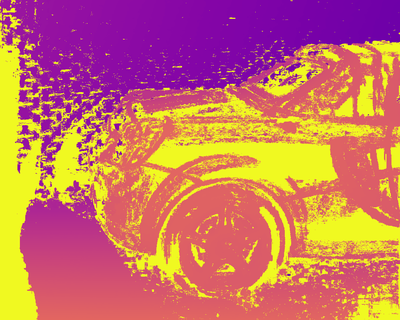} &
    \includegraphics[width=0.25\linewidth]{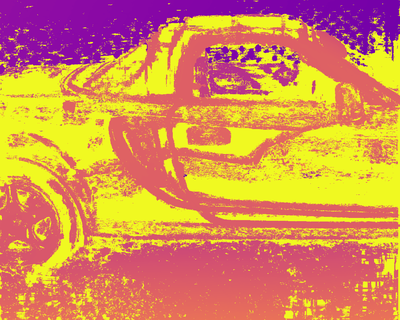} &
    \includegraphics[width=0.25\linewidth]{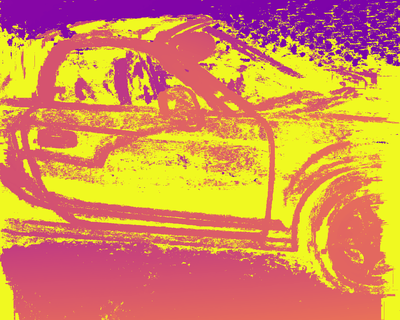}\\[-3.5pt]
    \includegraphics[width=0.25\linewidth]{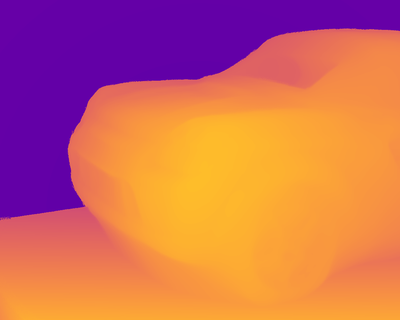} &
    \includegraphics[width=0.25\linewidth]{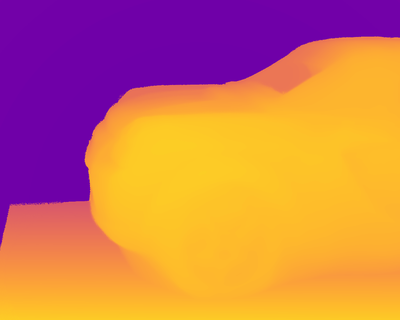} &
    \includegraphics[width=0.25\linewidth]{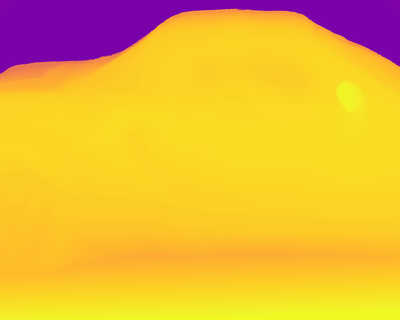} &
    \includegraphics[width=0.25\linewidth]{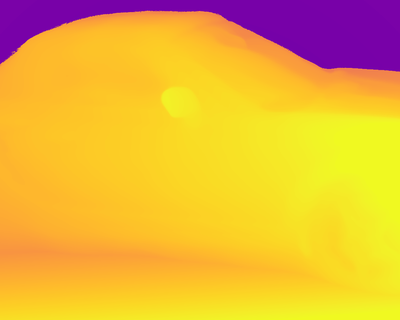}\\
    \end{tabular}
    \caption{Comparing ground-truth and predicted depth images from the unseen side of the car. Top: ground-truth sparse depth image. Bottom: our predicted depth image.}%
    \label{fig:ablation_depth}
\end{figure}

We also show the ground-truth point cloud for this scene in \cref{fig:ablation_pointcloud}. The ground-truth is very sparse and noisy, with incorrect geometry in numerous places. In contrast, our reconstruction is dense and reconstructs the geometry very well, despite seeing only half of the car. This can also be seen in \cref{fig:ablation_depth}, where the ground-truth depth map is plotted alongside our predicted depth map. When the symmetry components are removed from our model, the reconstruction is unable to recover any detail on the unseen side of the car. Instead, the smoothness prior of the SDF imposes a closure of the seen part of the car.

